\pdfoutput=1

\documentclass[11pt]{article}

\usepackage{acl}

\usepackage{times}
\usepackage{latexsym}

\usepackage[T1]{fontenc}

\usepackage[utf8]{inputenc}

\usepackage{microtype}

\usepackage{enumitem}
\setlist{nosep} 

\title{Evaluating  Deception Detection Model Robustness To Linguistic Variation}

\author{Maria Glenski, Ellyn Ayton, Robin Cosbey, Dustin Arendt, and Svitlana Volkova \\ 
  Pacific Northwest National Laboratory \\
  Richland, WA, USA \\
  {\tt first.last@pnnl.gov} \\}

\newcommand{\new}[1]{#1} 
  
\usepackage{xspace} 

\newcommand{\tr}{$Tr$\xspace} 
\newcommand{\trHalf}{$Tr^{50\%}$\xspace} 
\newcommand{\trFull}{$Tr^{'}$\xspace} 
\newcommand{\trHalfChar}{$Tr^{50\%}_{C}$\xspace} 
\newcommand{\trFullChar}{$Tr^{'}_{C}$\xspace} 
\newcommand{\trHalfWord}{$Tr^{50\%}_{W}$\xspace} 
\newcommand{\trHalfBert}{$Tr^{50\%}_{BERT}$\xspace} 
\newcommand{\trHalfElmo}{$Tr^{50\%}_{ELMo}$\xspace} 
\newcommand{\trHalfGlove}{$Tr^{50\%}_{GLoVe}$\xspace} 
\newcommand{\trFullWord}{$Tr^{'}_{W}$\xspace} 
\newcommand{\Ens}{$E$\xspace} 
\newcommand{\EnsChar}{$E_{C}$\xspace} 
\newcommand{\EnsWord}{$E_{W}$\xspace} 
\newcommand{\EnsCharWord}{$E_{C+W}$\xspace}

\newcommand{\te}{$Te$\xspace} 
\newcommand{\teFull}{$Te^{'}$\xspace} 
\newcommand{\teFullChar}{$Te^{'}_{C}$\xspace} 
\newcommand{\teFullWord}{$Te^{'}_{W}$\xspace} 
\newcommand{\teFullBert}{$Te^{'}_{BERT}$\xspace} 
\newcommand{\teFullElmo}{$Te^{'}_{ELMo}$\xspace} 
\newcommand{\teFullGlove}{$Te^{'}_{GLoVe}$\xspace} 
\newcommand{\teMixed}{$Mixed$\xspace} 
\newcommand{\teMixedDefinition}{$Te$ + $Te'_{C}$ + $Te'_{W}$\xspace} 
\newcommand{\teMixedBert}{$Mixed_{BERT}$\xspace} 
\newcommand{\teMixedElmo}{$Mixed_{ELMo}$\xspace} 
\newcommand{\teMixedGlove}{$Mixed_{GLoVe}$\xspace}

\usepackage[T1]{fontenc}  
\usepackage[utf8]{inputenc}

\newcommand{\ignore}[1]{}
\newcommand{\hide}[1]{}

\newcommand{\nop}[1]{} 
\newcommand{\eg}{\textit{e.g.,}\xspace}
\newcommand{\ie}{\textit{i.e.,}\xspace}

\usepackage{array}

\usepackage{multirow, multicol}

\usepackage{graphicx}  
\usepackage{subcaption,enumitem}

\usepackage{tikz}           
\usepackage{pgfplots}       
\usepgfplotslibrary{polar}  
\usepackage{xcolor,colortbl}
\usepackage{pgf-pie}        
\usepackage{pgfplotstable}  
\usepgfplotslibrary{fillbetween}    
\usetikzlibrary{pgfplots.statistics} 

\usepgfplotslibrary{groupplots}
\makeatletter
\newcommand\resetstackedplots{
\makeatletter
\pgfplots@stacked@isfirstplottrue
\makeatother
\addplot [forget plot,draw=none] coordinates{(1,0) (2,0) (3,0)};
}
\makeatother

\pgfplotsset{
        discard if not/.style 2 args={
            filter discard warning=false,
            x filter/.append code={
                \edef\tempa{\thisrow{#1}}
                \edef\tempb{#2}
                \ifx\tempa\tempb
                \else
                    \def\pgfmathresult{inf}
                \fi
            },
        },
    }

\usetikzlibrary{arrows}

 \colorlet{nonErrorBackground}{gray!5}
\colorlet{errors}{black!30}
\colorlet{Errors90}{black}
\colorlet{errors90}{black}
\colorlet{Cerrors}{cyan!30}
\colorlet{CErrors90}{cyan}
\colorlet{Werrors}{red!30}
\colorlet{WErrors90}{red}

\colorlet{BertErrors}{errors}
\colorlet{BertErrors90}{Errors90}
\colorlet{ElmoErrors}{errors}
\colorlet{ElmoErrors90}{Errors90}
\colorlet{GloveErrors}{errors}
\colorlet{GloveErrors90}{Errors90}

\colorlet{BertErrorsC}{Cerrors}
\colorlet{BertErrors90C}{CErrors90}
\colorlet{ElmoErrorsC}{Cerrors}
\colorlet{ElmoErrors90C}{CErrors90}
\colorlet{GloveErrorsC}{Cerrors}
\colorlet{GloveErrors90C}{CErrors90}

\colorlet{BertErrorsW}{Werrors}
\colorlet{BertErrors90W}{WErrors90}
\colorlet{ElmoErrorsW}{Werrors}
\colorlet{ElmoErrors90W}{WErrors90}
\colorlet{GloveErrorsW}{Werrors}
\colorlet{GloveErrors90W}{WErrors90}

\colorlet{mixed3way}{violet}
\colorlet{clean3way}{mixed3way!50!white}
\colorlet{mixed4way}{magenta}
\colorlet{clean4way}{mixed4way!50!white}

\colorlet{ratioHMneg}{blue}
\colorlet{ratioHMpos}{orange}

\colorlet{lowconf}{orange} 
\colorlet{mediumconf}{purple} 
\colorlet{highconf}{green}

\colorlet{TE}{black} 
\colorlet{TEc}{orange!75!black} 
\colorlet{TEw}{orange!50!white} 

\colorlet{TETEcTEw}{cyan}

\colorlet{BertTE}{blue!50!white}
\colorlet{BertTEc}{blue!50!white}
\colorlet{BertTEw}{blue!50!white}
\colorlet{BertTETEcTEw}{blue!75!black}

\colorlet{BertTE3}{black!50!white} 
\colorlet{BertTETEcTEw3}{black!80!white} 
\colorlet{BertTE4}{red!50!white} 
\colorlet{BertTETEcTEw4}{red!70!black}

\colorlet{BertTE}{black!80!white}
\definecolor{BertTEc}{RGB}{55,126,184}
\definecolor{BertTEw}{RGB}{77,175,74}
\colorlet{BertTETEcTEw}{black!50!white}

\colorlet{ElmoTE}{BertTE}
\colorlet{ElmoTEc}{BertTEc}
\colorlet{ElmoTEw}{BertTEw}
\colorlet{ElmoTETEcTEw}{BertTETEcTEw}

\colorlet{GloveTE}{BertTE}
\colorlet{GloveTEc}{BertTEc}
\colorlet{GloveTEw}{BertTEw}
\colorlet{GloveTETEcTEw}{BertTETEcTEw}

\tikzstyle{linestyleTE}= [solid, thick]
\tikzstyle{linestyleTEc}= [solid, thick]
\tikzstyle{linestyleTEw}= [solid, thick]
\tikzstyle{linestyleTETEcTEw}= [solid, thick]

\tikzstyle{linestyleBertTE}=[linestyleTE] 
\tikzstyle{linestyleBertTEc}= [linestyleTEc] 
\tikzstyle{linestyleBertTEw}= [linestyleTEw] 
\tikzstyle{linestyleBertTETEcTEw}=[linestyleTETEcTEw] 

\tikzstyle{linestyleElmoTE}= [linestyleTE]
\tikzstyle{linestyleElmoTEc}= [linestyleTEc]
\tikzstyle{linestyleElmoTEw}= [linestyleTEw]
\tikzstyle{linestyleElmoTETEcTEw}= [linestyleTETEcTEw]

\tikzstyle{linestyleGloveTE}= [linestyleTE]
\tikzstyle{linestyleGloveTEc}= [linestyleTEc]
\tikzstyle{linestyleGloveTEw}= [linestyleTEw]
\tikzstyle{linestyleGloveTETEcTEw}= [linestyleTETEcTEw]

\tikzstyle{lineplotMarkTE}= [mark=*, mark size=0.75, mark options={solid, thin}] 
\tikzstyle{lineplotMarkTEc}= [mark=*, mark size=0.75, mark options={solid,thin}]  
\tikzstyle{lineplotMarkTEw}= [mark=*, mark size=0.75, mark options={solid,thin}] 
\tikzstyle{lineplotMarkTETEcTEw}= [mark=*, mark size=0.75, mark options={solid,thin}]

\colorlet{radarTEc}{cyan}
\colorlet{radarTEw}{red}
\tikzstyle{radarplotMarkTEc}= [mark=*, mark options={solid,thick}]  
\tikzstyle{radarplotMarkTEw}= [mark=square*, mark options={solid,thick}]  
\tikzstyle{radarplotFillTEc}= [fill=radarTEc, fill opacity=0.2]  
\tikzstyle{radarplotFillTEw}= [fill=radarTEw, fill opacity=0.2]

\colorlet{radarTE}{black}
\colorlet{radarTEmixed}{purple}
\tikzstyle{radarplotMarkTEc}= [mark=x, mark options={solid,thick}]  
\tikzstyle{radarplotMarkTEw}= [mark=diamond*, mark options={solid,thick}]  
\tikzstyle{radarplotFillTE}= [fill=radarTE, fill opacity=0.2]  
\tikzstyle{radarplotFillTEmixed}= [fill=radarTEmixed, fill opacity=0.2]

\begin{document}
\maketitle

\begin{abstract}
With the increasing use of machine-learning driven algorithmic judgements, it is critical to develop models that are robust to evolving or manipulated inputs. We propose an extensive analysis of model robustness against linguistic variation in the setting of \textit{deceptive news detection}, an important task in the context of misinformation spread online.
We consider two prediction tasks and compare three state-of-the-art embeddings to highlight consistent trends in model performance, high confidence misclassifications, and high impact failures. 
By measuring the effectiveness of adversarial defense strategies and evaluating model susceptibility to adversarial attacks using character- and word-perturbed text, we find that character or mixed ensemble models are the most effective defenses and that character perturbation-based attack tactics are more successful.
\end{abstract}

\noindent


\section{Introduction}

Over two-thirds of US adults get their news from social media, but over half (57\%) ``expect the news they see on social media to be largely inaccurate''~\cite{shearer2018}. 
A 2020 Reuters Institute global news survey found a similar trend with 56\% of respondents concerned with misinformation in online news~\cite{reuters2020digital}. 
There are online and offline impacts from the spread of misinformation or deceptive news stories within online communities. However, the rate at which new content is submitted to social media platforms is a significant obstacle for approaches that require manual identification, annotation, or intervention.  In recent efforts, evaluation has focused on aggregate performance metrics on test sets often collected from social media platforms like Twitter, Facebook, or Reddit~\cite{rubin2016fake,mitra2017parsimonious,wang2017liar} but these platforms are not representative in regards to user demographics or topics of discussion. Further, aggregate performance metrics are not sufficient to provide insight on generalizable performance. 

When we consider the identification of deceptive news online --- where humans often disagree on or challenge the judgements of others~\cite{karduni2018can,karduni2019vulnerable,ott2011finding} --- we need more rigorous evaluations of model decisions, with a focus on expected performance across varied or manipulated inputs. 
Our work examining reliability of performance when faced with linguistic variations is a step towards comprehensively understanding model robustness that may highlight inequalities in cases of failure. 
Although machine learning models are often leveraged for their ability to tackle rapid response at scale, it is critical to understand nuanced model biases and the significant downstream consequences of model decisions on users.  

A known gap exists in our understanding of underlying machine learning decision-making processes, particularly with deep learning ``black-box'' models.
 The use of traditional, aggregate metrics for model performance, such as accuracy or F1 score, are not sufficient in pursuit of this understanding. 
We argue that evaluations need to explicitly measure the extent to which model performance is affected by data with a varied topic distribution.  
Evaluations highlighting when models are correct, which examples can provide explanations, and clarification or reasoning for why a user should trust a given model  are well-aligned with recent themes in research on machine learning interpretability, trust, fairness, accountability, and reliability~\cite{lipton2016mythos,doshi2017towards,hohman2018visual}.
 
In this paper, we perform an adversarial model evaluation across two multimodal deception prediction tasks to identify which defensive strategies are most successful across a variety of attacks. 
Our main contribution is a framework of analysis for model robustness across variations in linguistic signals and representations that may be encountered in real-world applications of digital deception models (\eg natural linguistic differences, evolving tactics from deceptive adversaries to evade detection).  In particular, we present evaluations on the susceptibility of widely used text embeddings to naive adversarial attacks, which types of text perturbations lead to the most high-confident errors, and to what extent our findings are task specific.  
The perturbed text emulates real examples of linguistic variations, e.g. non-native speakers, spelling mistakes, or shortened online speech. 
Our evaluations reveal how models react to perturbed text which we argue is a likely occurrence when deployed in a real-world setting.

\section{Related Work}

With the increasing concern for the impact of misinformation and deceptive news content online, many studies have explored or developed models that detect such news. 
Recent efforts focus on identifying a spectrum of deception: from binary classification of content as suspicious and trustworthy \cite{volkova2017separating} to a more fine-grained separation within deceptive classes (\eg propaganda, hoax, satire) \cite{rashkin2017truth}.
Additional work has explored the behavior of malicious users and bots \cite{glenski2018humans,kumar2017army,kumar2018rev2} and spread patterns of misinformation or rumors \cite{kwon2017rumor,vosoughi2018spread} to aid in classification tasks. 
Strong evidence suggests that enriched features such as images, temporal and structural attributes, and linguistic features boost model performance over dependence on textual characteristics alone \cite{wang2017liar,qazvinian2011rumor,kwon2013prominent}.  
The need for effective, trustworthy, and interpretable detection models is a vital concern and must be an essential requirement for models where decisions or recommendations can significantly affect end users.  

A variety of deep learning architectures applied to deception detection tasks include convolutional neural networks (CNNs)
~\cite{ajao2018fake,wang2017liar,volkova2017separating}, long short-term memory (LSTM) models~\cite{chen2018unsupervised,rath2017retweet,zubiaga2018discourse,zhang2019reply}  
, and LSTM variants with attention mechanisms~\cite{guo2018rumor,li2019rumor}.
Architecture and other aspects of neural network design typically depend on the classification task and require specialized hyperparameter tuning.
In order to provide a fair comparison of model evaluations across tasks and for the purpose of consistency across experiments, we implement a multimodal LSTM model similar to recent work. 
Our approach allows for more accurate comparisons of factors related to adversarial susceptibility across classification tasks. 
Developing novel state-of-the-art models for deception detection or comparing multiple architectures is beyond the scope of this paper. 
 
Although popularly used across many domains,  
deep learning systems can be extremely brittle when evaluated on examples 
outside of the training data distribution~\cite{goodfellow2014explaining,moosavi2017universal,fawzi2018analysis}.  
Nguyen \textit{et al}~\shortcite{nguyen2015deep} have shown that small perturbations in input data can cause highly probable misclassifications. 
Further research demonstrates additional attacks that make neural networks more susceptible to adversaries such as locally trained DNNs to crafted adversarial inputs \cite{DBLP:journals/corr/PapernotMGJCS16,papernot2016limitations} and gradient-based attacks \cite{biggio2013evasion}.  
To counteract these offensive strategies, proposed  methods of defense include augmented training data with adversarial examples~\cite{tramer2017ensemble},   
training a separate model to distinguish genuine data from malicious data \cite{metzen2017detecting},  
and implementing a defensive distillation mechanism to increase a model's resiliency to data poisoning \cite{papernot2016distillation}. 
However, as defense strategies are created, new attacks are continually developed to circumvent them~\cite{carlini2017towards}. 
While there is a focus on image perturbations and related attacks, textual data is similarly vulnerable to such strategies~\cite{gao2018black,samanta2017towards,liang2017deep}). 
The susceptibility of deception detection models to text-based adversarial attacks as well as the effectiveness of defense strategies have not been extensively evaluated.

\section{Methodology}
In this section, we introduce our detection tasks, models, and evaluation methods.
We randomly perturb words or characters with their nearest neighbors to mimic a low-effort adversarial attack (\eg replacing words with synonyms) as opposed to methods that assume an adversary has technical expertise or require sophisticated augmentations (\eg gradient-based algorithms). We argue that robustness against these low-effort attacks is a necessary first step towards trustworthy models; these attacks are reflective of natural or unintentional variations (\eg misspellings, non-native speaker discussions) as well as sophisticated strategies.

\subsection{Deception Detection Tasks} 
We apply a comprehensive evaluation of model robustness and susceptibility to two classification tasks\footnote{Although we chose to use these two tasks, our framework is task-agnostic and can be applied to any classification task.}: 3-way (trustworthy, propaganda, disinformation) and 4-way (clickbait, hoax, satire, conspiracy). 
Including both allows us to compare defense and attack strategies across models at varied levels of deception and evaluate method generalizability.

The 3-way task includes two extreme deceptive classes, propaganda and disinformation, and seeks to differentiate them from ``trustworthy'' sources~\cite{derakhshan2017information}.
Due to the stronger intent to deceive of these classes, we expect a model to distinguish trustworthy news more easily and expect more confusion when classifying news as either propaganda or disinformation.
Misclassifications of these as trustworthy will have a greater negative impact.
To better identify high-impact errors, we collapse disinformation and propaganda into a single class as part of a binary sub-task separating trustworthy from deceptive. 
  
The 4-way task centers lesser deceptive content, the classes included have a lower intent to deceive and are more difficult to distinguish from one another. 
For instance, satirical news sites produce humorous content or social commentary rather than deliberately false information and   
have a low intent to deceive audiences~\cite{fletcher2017people}. 
Because of this inherent  
difference from  
the other deceptive news types, we include a binary sub-task separating satire from the remaining classes.  

\subsection{Data Collection and Annotation}
Models were trained and tested on Twitter API data.
Our corpus comprises English retweets with images from official news media Twitter accounts.  
Class labels are based on ``verified'' news sources and a public list of sources annotated along the spectrum of deceptive content~\cite{volkova2017separating}\footnote{\url{www.cs.jhu.edu/~svitlana/data/SuspiciousNewsAccountList.tsv}; \url{www.cs.jhu.edu/~svitlana/data/VerifiedNewsAccountList.tsv}
} from 2016. Thus, we limit our corpus to that 12 month period of activity.
The 3-way and 4-way task data consist of 54.5k and 2.5k tweets.   
Although there are limits to source-level annotations (\eg tweets of different deceptive classes shared from a single source), we advocate for focus on news
sources rather than individual stories, similar to previous work ~\cite{vosoughi2018spread,lazer2018science}. 
We posit the definitive element of deception to be the intent and tactics of the source.

\subsection{Multimodal Deception Detection Models}

We clean the tweet text by lowercasing and removing punctuation, mentions, hashtags, and URLs. 
We encode biased and subjective language as frequency vectors constructed from LIWC~\cite{pennebaker2001linguistic} and several lexical dictionaries such as hedges and factives~\cite{recasens2013linguistic} which are often used for text classification~\cite{rashkin2017truth,shu2019beyond}.  
 
We implement a two-branch architecture\footnote{Parameters selected by a random search: Adam optimizer, $10^{-6}$ learning rate, 0.2 drop out, and 10 training epochs.} 
that leverages text, lexical features, and images. 
The text branch consists of a pre-trained text embedding layer, an LSTM layer, and a fully connected layer. The output is concatenated to the lexical feature vector before being passed to another fully connected layer.
In the second branch, we pass the image vector through a fully connected, two layer network. 
The combined text embeddings and lexical features are concatenated with the processed image representation which is then fed to a fully connected network for classification. 
Our chosen architecture resembles current systems in deployment and allows us to complete complex analyses. 

\subsection{Model Evaluation Methods}
We perform a comprehensive evaluation for both tasks over {\it embeddings}, {\it defenses}, and {\it attacks}. This section describes our text perturbation methods and our defense and attack frameworks.

\subsubsection{Varying Text Representations} 
We consider three embedding techniques that have shown state-of-the-art performance on several NLP tasks: 
GLoVe~\cite{pennington2014glove}, 
ELMo~\cite{peters2018elmo}, 
and BERT~\cite{devlin2019bert}. 
We recognize that each embedding method was trained on separate data\footnote{We use GloVe (Twitter 27B), ELMo (\url{tfhub.dev/google/elmo/2}), and BERT (\url{github.com/huggingface/transformers})}, under different conditions, and produces various sized vectors. 
Thus, we fine-tune the embedding layer during training.
 
\begin{figure}
    \centering
    \includegraphics[scale=.56]{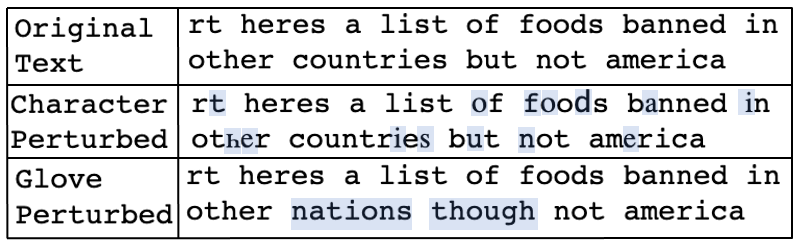} 
     \vspace{-0.5\baselineskip}
    \caption{Examples of adversarial perturbations.}
    \label{perturb_text}
\end{figure}

\subsubsection{Linguistic Variation}
We examine how changes to text input affect model performance using character and word perturbations and focus on the impact of naive linguistic variations in text. 
For character-level perturbations, we randomly replace 25\% of characters in each tweet with a Unicode character that is indistinguishable from the original to a human (as shown in Figure~\ref{perturb_text}). 
This approach, known in computer security as a homograph attack or script spoofing, has been investigated to identify phishing or spam \cite{fu2006safeguard,fu2006methodology,liu2007fighting} but has not been applied in the NLP domain to our knowledge.
For word-level perturbations we randomly replace 25\% of words with a nearest neighbor in the each embedding space using Annoy\footnote{\url{https://github.com/spotify/annoy}}. 

\subsubsection{Defense Viewpoint}
To evaluate the efficacy of common defenses to guard against adversarial attacks -- augmenting the training data -- we perturb our training set (\tr) to varying degrees \new{using each linguistic variation strategy}. We compare the following defenses:

\begin{itemize}[topsep=1pt,itemsep=1pt,partopsep=0pt, parsep=0pt] 
	\item \tr: train with original examples; 
	\item \trHalf: train with half of the examples perturbed; 
	\item \trFull: train with all examples perturbed; 
	\item Ensemble (\Ens): majority vote of ensemble of models trained on \tr, \trHalf, and \trFull.  
\end{itemize}
Each defense has been perturbed for each embedding type. For example, we train our models using four variations of \trHalf:  \trHalfChar (50\% of examples perturbed using the character-level attack), and three \trHalfWord defenses with 50\% of the examples perturbed using the word-level attack (\trHalfBert, \trHalfElmo, \trHalfGlove). 

We use three sets of ensembles: (1) \EnsChar, an ensemble of models trained with \tr, \trHalfChar, and \trFullChar, (2) \EnsWord, an ensemble of models trained with \tr, \trHalfWord, and \trFullWord, and (3) \EnsCharWord, an ensemble of five models trained on \tr, \trHalfChar, \trFullChar, \trHalfWord, and \trFullWord. 
Higher confidence predictions are used in the case of ties. 

We test the performance of models trained \new{using each defense} on fully perturbed (\teFull) and the original, unperturbed test data (\te).
Ideally, we want models to perform well on both so we also consider three \teMixed test sets (\teMixedDefinition), one for each \teFullWord (\teMixedBert, \teMixedElmo, and \teMixedGlove). 
 
\subsubsection{Attack Viewpoint}
We also evaluate the impact of the linguistic perturbations as adversarial attack strategies. The attack test sets were perturbed similarly to the train sets:  
\begin{itemize}[topsep=1pt,itemsep=1pt,partopsep=0pt, parsep=0pt] 
	\item 
	\te: 
	original examples (no attack); 
	\item 
	\teFullChar: 
	 all examples perturbed 
	 (char-level); 
	\item 
	\teFullWord: 
	 all examples perturbed 
	 (word-level). 
\end{itemize}
 
As with the defense viewpoint, we have four sets of the \teFull test data used to evaluate each attack condition: \teFullChar,  \teFullBert, \teFullElmo, \teFullGlove.  

\ignore{
\vspace{-\baselineskip}
\begin{itemize} 
	\item{\bf Character-level} We randomly replaced 25\% of characters with a Unicode character that is indistinguishable to the original to a human\footnote{A few characters like \textit{Cyrillic Small Letter Shha} are distinguishable when viewed with certain fonts.} using a publicly available  list\footnote{\url{https://www.unicode.org/Public/security/12.1.0/intentional.txt}} of Unicode characters. Our approach 
	is known in computer security as a homograph attack, Unicode attack, or script spoofing. These types of deliberate attacks have been investigated to identify phishing and spam \cite{fu2006safeguard,fu2006methodology,liu2007fighting} but have not been applied in the NLP domain to our knowledge.
	
	\vspace{-0.5\baselineskip}
	\item{\bf Word-level} We randomly replace 25\% of words with a nearest neighbor in the corresponding (GloVe, BERT, or ELMo) embedding space using Annoy\footnote{\url{https://github.com/spotify/annoy}} to efficiently perform the approximate nearest neighbor search.
\end{itemize}
}

\subsubsection{High Confidence And High Impact}  
For researchers and end-users to establish trust in the models they develop or use, it is essential to understand the circumstances in which a model would make a highly confident misclassification. 
Inherently, model confidence measures the certainty of a prediction and quantifies the expertise and stability of a model. 
We closely examine instances in which our models have incorrectly predicted the class of a tweet with high confidence (greater than 90\%) to identify potential weaknesses of the models.  
 
Traditional performance metrics (F1 score, precision, recall) treat misclassifications with high confidence and low confidence alike. 
While overall error is an important measure, a model with a slightly higher overall failure rate but lower confidence may result in a better ``worst case'' outcome if appropriately incorporated in a semi-automated or human-in-the-loop deployment strategy that considers the uncertainty of predictions or recommendations via the model confidence before taking action. 

We also examine high impact errors using the binary sub-tasks for each classification task as described above.
In this analysis, we identify how often models make significant errors. 
For example, mistaking a post labeled as disinformation for trustworthy (an opposite class) rather than propaganda (a similar class).

\section{Experimental Results} 
 
In this section, we detail our results when evaluating different combinations of adversarial defenses and attacks. 
In order to produce a holistic evaluation of model susceptibility, we examine defenses and attacks separately.
Although we consider the same model behavior, each position can highly impact the interpretation of the findings and key takeaways. 
We also want to understand model misclassifications, including those with high model confidence and those that can have a greater negative effect in practice which we accomplish with our high confidence and high impact analyses. 

\subsection{Defense Viewpoint}
We compare results from the models trained on data with varying degrees of perturbation to understand which models provide the most effective defenses. 
We define success in the defender case as the lowest error rate across  
a variety of test data including original (\te), perturbed (\teFull), and combinations of original and perturbed samples (\teMixed). We start by presenting the \textit{relative difference} in error rates which is the percentage increase or decrease in the error rate of the perturbed ($\hat{Te}'$) and original ($\hat{Te}$) test data. Relative difference is defined as:
\begin{equation}
    \Delta Te'_{x} = \frac{\hat{Te}'_{x}-\hat{Te}}{\hat{Te}}
\end{equation}
where $x$ represents perturbation type (char or word).
Relative difference results are shown in Table~\ref{tab:relative_error_delta_on_perturbed}.

\begin{table}[t!]
	\centering
	\small 
	
\centering
\footnotesize
\setlength\tabcolsep{3pt} 
\begin{tabular}{lrrr@{\hskip .4cm}rrr} 
	{\footnotesize \textit{3-way}}
	&
	\multicolumn{3}{c}{  \bf Character ($\Delta Te'_{C}$)}
	&
	\multicolumn{3}{c}{  \bf Word ($\Delta Te'_{W}$)}
	\\ 
	
	  \textbf{Defense}
	& \footnotesize \it  BERT 
	& \footnotesize \it  ELMo 
	& \footnotesize \it  GloVe 
	
	& \footnotesize \it  BERT 
	& \footnotesize \it  ELMo 
	& \footnotesize \it  GloVe 
	\\
	 \tr 
	&  +36\% \cellcolor{ratioHMpos!36}
	&  +34\%  \cellcolor{ratioHMpos!34}
	&  +33\%  \cellcolor{ratioHMpos!33}
	&  +37\% \cellcolor{ratioHMpos!37}
	&  +37\% \cellcolor{ratioHMpos!37}
	&  +38\% \cellcolor{ratioHMpos!38}
	
	\\
	 \trHalf 
	&  +2\% \cellcolor{ratioHMpos!2}
	&  -2\% \cellcolor{ratioHMneg!2}
	&  -2\% \cellcolor{ratioHMneg!2}
	&  -1\% \cellcolor{ratioHMneg!1}
	&  -3\% \cellcolor{ratioHMneg!3}
	&  -3\% \cellcolor{ratioHMneg!3}
	
	\\
	 \trFull 
	&  +1\% \cellcolor{ratioHMpos!1}
	&  -1\% \cellcolor{ratioHMneg!1}
	&  -1\% \cellcolor{ratioHMneg!1}
	&  -6\% \cellcolor{ratioHMneg!6}
	&  -21\% \cellcolor{ratioHMneg!21}
	&  -5\% \cellcolor{ratioHMneg!5}
	
	\\
	 \EnsChar 
	&  +2\% \cellcolor{ratioHMpos!2}
	&  +4\% \cellcolor{ratioHMpos!4}
	&  +7\% \cellcolor{ratioHMpos!7}
	&  -5\% \cellcolor{ratioHMneg!5}
	&  -8\% \cellcolor{ratioHMneg!8}
	&  -0\% \cellcolor{ratioHMneg!0}
	
	\\
	 \EnsWord 
	&  +14\% \cellcolor{ratioHMpos!14}
	&  +12\% \cellcolor{ratioHMpos!12}
	&  +15\% \cellcolor{ratioHMpos!15}
	&  +11\% \cellcolor{ratioHMpos!11}
	&  +21\% \cellcolor{ratioHMpos!21}
	&  +17\% \cellcolor{ratioHMpos!17}
	
	\\
	 \EnsCharWord 
	&  +10\% \cellcolor{ratioHMpos!10}
	&  +7\% \cellcolor{ratioHMpos!7}
	&  +7\% \cellcolor{ratioHMpos!7}
	&  +7\% \cellcolor{ratioHMpos!7}
	&  +9\% \cellcolor{ratioHMpos!9}
	&  +3\% \cellcolor{ratioHMpos!3}
	
	\\

	\\[-0.75em]

	{\footnotesize \textit{4-way}}
	&
	\multicolumn{3}{c}{  \bf Character ($\Delta Te'_{C}$)}
	&
	\multicolumn{3}{c}{  \bf Word ($\Delta Te'_{W}$)}
	\\

	 \textbf{Defense}  
	& \footnotesize \it  BERT 
	& \footnotesize \it  ELMo 
	& \footnotesize \it  GloVe 
	
	& \footnotesize \it  BERT 
	& \footnotesize \it  ELMo 
	& \footnotesize \it  GloVe 
	\\
	 \tr 
	&  +14\%  \cellcolor{ratioHMpos!14}
	&  +52\%  \cellcolor{ratioHMpos!52}
	&  +0\%  \cellcolor{ratioHMpos!0}
	&  +5\% \cellcolor{ratioHMpos!5}
	&  +53\% \cellcolor{ratioHMpos!53}
	&  +6\% \cellcolor{ratioHMpos!6}
	\\
	 \trHalf 
	&  +32\% \cellcolor{ratioHMpos!32}
	&  +31\% \cellcolor{ratioHMpos!31}
	&  +36\% \cellcolor{ratioHMpos!36} 
	&  +16\% \cellcolor{ratioHMpos!16}
	&  +10\% \cellcolor{ratioHMpos!10}
	&  +16\% \cellcolor{ratioHMpos!16}
	\\
	 \trFull 
	&  +30\% \cellcolor{ratioHMpos!30}
	&  +30\% \cellcolor{ratioHMpos!30}
	&  +32\% \cellcolor{ratioHMpos!32}
	&  -7\% \cellcolor{ratioHMneg!7}
	&  -3\% \cellcolor{ratioHMneg!3}
	&  -5\% \cellcolor{ratioHMneg!5}
	\\
	 \EnsChar 
	&  +11\% \cellcolor{ratioHMpos!11}
	&  +15\% \cellcolor{ratioHMpos!15}
	&  +2\% \cellcolor{ratioHMpos!2}
	&  +13\% \cellcolor{ratioHMpos!13}
	&  +17\% \cellcolor{ratioHMpos!17}
	&  +10\% \cellcolor{ratioHMpos!10}
	\\
	 \EnsWord 
	&  +28\% \cellcolor{ratioHMpos!28}
	&  +48\% \cellcolor{ratioHMpos!48}
	&  +21\% \cellcolor{ratioHMpos!21} 
	&  +8\% \cellcolor{ratioHMpos!8}
	&  +25\% \cellcolor{ratioHMpos!25}
	&  +6\% \cellcolor{ratioHMpos!6}
	\\
	 \EnsCharWord 
	&  +17\% \cellcolor{ratioHMpos!17}
	&  +21\% \cellcolor{ratioHMpos!21}
	&  +22\% \cellcolor{ratioHMpos!22}
	&  +17\% \cellcolor{ratioHMpos!17}
	&  +9\% \cellcolor{ratioHMpos!9}
	&  +15\% \cellcolor{ratioHMpos!15}
	\\
\end{tabular}
 
\vspace{-0.25\baselineskip} 
\setlength\minrowclearance{0pt}
\vspace{0.5\baselineskip} 
\begin{tabular}{lllllllll}
	\multicolumn{3}{c}{\footnotesize Fewer errors on $Te'$}
	&
	&\multicolumn{3}{c}{\footnotesize More errors on $Te'$}\\
	\cellcolor{ratioHMneg!60} ~ \textcolor{ratioHMneg!60}{XX}
	&   
	\cellcolor{ratioHMneg!40} ~\textcolor{ratioHMneg!40}{XX}
	&   
	\cellcolor{ratioHMneg!20} ~\textcolor{ratioHMneg!20}{XX}
	&  
	&
	\cellcolor{ratioHMpos!20} ~\textcolor{ratioHMpos!20}{XX}
	&
	\cellcolor{ratioHMpos!40} ~\textcolor{ratioHMpos!40}{XX}
	&
	\cellcolor{ratioHMpos!60} ~ \textcolor{ratioHMpos!60}{XX}\\

\end{tabular} 
 
    \vspace{-0.5\baselineskip}
	\caption{Relative difference in error rate for each task's perturbed test data (\teFull) compared to original \te.  
	} 
	\label{tab:relative_error_delta_on_perturbed}
\end{table}

With the 3-way task, defenses across embeddings and test data appear effective and achieve low relative percent differences with the exception of models trained with the original examples (\tr). 
The \tr defense is ineffective against both the character- and word-perturbed text (\teFullChar and \teFullWord).
Intuitively, this could be seen as an "out of domain" data attack where the perturbed test set has significantly changed the original distribution such that a model not trained on perturbed data is more susceptible to errors. 
The \EnsChar models have a lower relative difference in errors on \teFullWord than on \teFullChar across all three embeddings used for text representations. 
Thus, an ensemble of models overcomes the setback of out of domain data.
 
On \teFullChar data, we observe similar relative errors between the three embeddings for all defense types;  however, the performance on \teFullWord is much more varied with the largest change seen from the ELMo embeddings,  -1\% relative difference from the \trFull model on \teFullChar and -21\% relative difference from the same defense on \teFullWord.
We only see consistent behavior with the \trHalf defense when tested on \te and \teFull across embedding strategies and attack perturbations.
A model trained on data containing 50\% clean and 50\% perturbed samples performs almost equally on the clean and perturbed test sets and exhibits less than a 5\% difference in errors between the test sets for all embeddings.

Dissimilarly, the 4-way task defenses display higher relative differences in errors on \teFullChar and \teFullWord with the exception of the \trFull defense. Under the \teFullWord attack, \trFull is the only defense to achieve fewer errors on the perturbed test set. 
We also see more variation in the relative errors across embeddings for the same defenses. For instance, with BERT, the \tr model defending against the \teFullChar attack has a 14\% relative error difference while the equivalent ELMo and GloVe models have 52\% and 0\% relative error differences, respectively. This trend appears across defenses and in some cases highlights the ineffectiveness of these defenses.

With both tasks, there are fewer errors on \teFullWord using \trFull as the defense, regardless of embedding type, specifically 21\% fewer errors on the 3-way task and 3\% fewer errors on the 4-way task with ELMo embeddings. 
Although the results on the tasks look dissimilar in terms of ``best generalizability" (\ie show good performance on both \te and \teFull), we see that character-based ensemble models exhibit the most consistent defense across tasks. 
The ability to have a single model (\EnsChar) perform uniformly well across tasks outweighs the slight performance increase with individualized models per task. 
The ensemble defenses that leverage character-based defenses (\EnsChar or \EnsCharWord) are more generalizable to novel test data which is beneficial when considering real-world data.

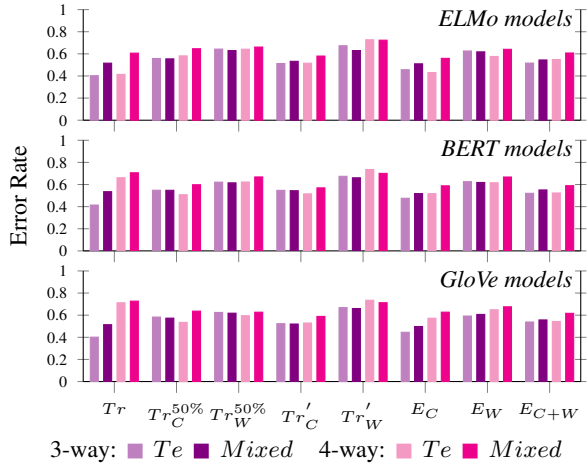
\begin{figure}[t!]
    \centering
    \centering

\begin{tikzpicture}
\begin{axis}[ 
    title ={\footnotesize \textit{ELMo models}}, 
    title style = {yshift=-.05in, 
    xshift=1.3in, 
    anchor=east}, 
    axis x line*=left, 
	ylabel = \textcolor{white}{Error Rate}, 
	ylabel style = {yshift=-.15in,
	font=\footnotesize}, 
	height = 1.2in,  
	width = 3.2in, 
	xtick align = center, ytick align = center, 
	title style = {yshift = -.075in,align = center}, 
	xtick pos = left,    
	ymin=0,ymax=1, 
	xtick={0,1,2,3,4,5,6,7},
	xmin=-0.5,xmax=7.5, 
	xticklabels={\tr,\trHalfChar,\trHalfWord,\trFullChar,\trFullWord,\EnsChar,\EnsWord,\EnsCharWord}, 
	xticklabels={,,,,,,,}, 
	xticklabel style= { 
	align=center, anchor= north}, 
	ybar,
	bar width=3pt,
	ytick={0,0.2,0.4,0.6,0.8,1},
	xticklabel style={font=\tiny},
	yticklabel style={font=\tiny},
	yticklabels={ 0, 0.2, 0.4, 0.6, 0.8, 1}, 
	]
	
	 \addplot+[clean3way,fill=clean3way, 
	 ybar] table [col sep=comma,y=elmoTE, x=x]{figs/tikz_data/fig2_3way_all_models_data.csv}; 
	 
	 \addplot+[mixed3way,fill=mixed3way,
	 ybar] table [col sep=comma,y=elmoTETEcTEw, x=x]{figs/tikz_data/fig2_3way_all_models_data.csv};

	 \addplot+[clean4way,fill=clean4way,
	 ybar,  mark size=3pt] table [col sep=comma,y=elmoTE, x=x]{figs/tikz_data/fig2_4way_all_models_data.csv};
	 
	 \addplot+[mixed4way,fill=mixed4way,
	 ybar] table [col sep=comma,y=elmoTETEcTEw, x=x]{figs/tikz_data/fig2_4way_all_models_data.csv};

\end{axis}
\end{tikzpicture} 
 
 \vspace{-0.8\baselineskip} 
\begin{tikzpicture}
\begin{axis}[ 
    title ={\footnotesize \textit{BERT models}}, 
    title style = {yshift=-.05in, 
    xshift=1.3in, 
    anchor=east}, 
    axis x line*=left,
	ylabel = Error Rate,
	ylabel style = {yshift=-.15in,
	font=\footnotesize},
	height = 1.2in,
	width = 3.2in,
	xtick align = center, ytick align = center, 
	title style = {yshift = -.075in,align = center}, 
	xtick pos = left,   
	ymin=0,ymax=1,
	xtick={0,1,2,3,4,5,6,7},
	xmin=-0.5,xmax=7.5,
	xticklabels={\tr,\trHalfChar,\trHalfWord,\trFullChar,\trFullWord,\EnsChar,\EnsWord,\EnsCharWord}, 
	xticklabels={,,,,,,,},  
	xticklabel style= {
	align=center, anchor= north},
	ybar,
	bar width=3pt,
	ytick={0,0.2,0.4,0.6,0.8,1},
	xticklabel style={font=\tiny},
	yticklabel style={font=\tiny},
	yticklabels={ 0, 0.2, 0.4, 0.6, 0.8, 1},
	]
	
	 \addplot+[clean3way,fill=clean3way, 
	 ybar] table [col sep=comma,y=bertTE, x=x]{figs/tikz_data/fig2_3way_all_models_data.csv}; 
	 
	 \addplot+[mixed3way,fill=mixed3way,
	 ybar] table [col sep=comma,y=bertTETEcTEw, x=x]{figs/tikz_data/fig2_3way_all_models_data.csv};

	 \addplot+[clean4way,fill=clean4way,
	 ybar,  mark size=3pt] table [col sep=comma,y=bertTE, x=x]{figs/tikz_data/fig2_4way_all_models_data.csv};
	 
	 \addplot+[mixed4way,fill=mixed4way,
	 ybar] table [col sep=comma,y=bertTETEcTEw, x=x]{figs/tikz_data/fig2_4way_all_models_data.csv};

\end{axis}
\end{tikzpicture} 

 \vspace{-0.8\baselineskip} 
\begin{tikzpicture}
\begin{axis}[ 
    title ={\footnotesize \textit{GloVe models}}, 
    title style = {yshift=-.05in, 
    xshift=1.3in, 
    anchor=east}, 
    axis x line*=left,
	xlabel style={yshift=-0.5\baselineskip,
	font=\footnotesize},
	ylabel = \textcolor{white}{Error Rate},
	ylabel style = {yshift=-.15in,
	font=\footnotesize},
	height = 1.2in,
	width = 3.2in,
	xtick align = center, ytick align = center, 
	title style = {yshift = -.075in,align = center}, 
	xtick pos = left,   
	ymin=0,ymax=1,
	xtick={0,1,2,3,4,5,6,7},
	xmin=-0.5,xmax=7.5,
	xticklabels={\tr,\trHalfChar,\trHalfWord,\trFullChar,\trFullWord,\EnsChar,\EnsWord,\EnsCharWord}, 
	xticklabel style= {
	align=center, anchor= north},
	ybar,
	bar width=3pt,
	ytick={0,0.2,0.4,0.6,0.8,1},
	xticklabel style={font=\tiny},
	yticklabel style={font=\tiny},
	yticklabels={ 0, 0.2, 0.4, 0.6, 0.8, 1},
	]
	
	 \addplot+[clean3way,fill=clean3way, 
	 ybar] table [col sep=comma,y=gloveTE, x=x]{figs/tikz_data/fig2_3way_all_models_data.csv}; 
	 
	 \addplot+[mixed3way,fill=mixed3way,
	 ybar] table [col sep=comma,y=gloveTETEcTEw, x=x]{figs/tikz_data/fig2_3way_all_models_data.csv};

	 \addplot+[clean4way,fill=clean4way,
	 ybar,  mark size=3pt] table [col sep=comma,y=gloveTE, x=x]{figs/tikz_data/fig2_4way_all_models_data.csv};
	 
	 \addplot+[mixed4way,fill=mixed4way,
	 ybar] table [col sep=comma,y=gloveTETEcTEw, x=x]{figs/tikz_data/fig2_4way_all_models_data.csv};

\end{axis}
\end{tikzpicture} 
  
\vspace{-0.5\baselineskip}

\begin{tikzpicture}
\begin{axis}[
hide axis,
height = .75in,width=1in,
xmin = 0, xmax = 50, ymin = 0, ymax = 0.4,
legend cell align = {left}, legend columns= 6,
legend style = {column sep=0.04in,
font=\footnotesize,
draw=none, at = {(0,0.75)}},]

\addlegendimage{empty legend}
\addlegendentry{ 3-way:}; 
\addlegendimage{only marks, mark = square*, clean3way, fill=clean3way, ultra thick}
\addlegendentry{$Te$}; 
\addlegendimage{only marks, mark = square*, mixed3way, fill=mixed3way, ultra thick}
\addlegendentry{$Mixed$};

\addlegendimage{empty legend}
\addlegendentry{4-way:}; 
\addlegendimage{only marks, mark = square*, clean4way, fill=clean4way, ultra thick}
\addlegendentry{$Te$}; 
\addlegendimage{only marks, mark = square*, mixed4way, fill=mixed4way, ultra thick}
\addlegendentry{$Mixed$};
  
\addlegendimage{only marks, mark = square*, white, fill=white, ultra thick}
\addlegendentry{\textcolor{white}{...}}; 

\end{axis}
\end{tikzpicture}  
\vspace{-2\baselineskip}

	\caption{
		Defense effectiveness illustrated by error rate as a function of defense strategy for each model when tested on \te or \teMixed (\te + \teFullChar + \teFullWord) data.}
	\label{fig:line_plots_defense_effectiveness}
\end{figure}

Performance on a variety of test data alone does not indicate the best defense. 
If a given defense performs similarly across datasets, it may simply perform equally poorly. 
Pairing additional analysis shown in Figure~\ref{fig:line_plots_defense_effectiveness} with generalizability results highlighted in Table~\ref{tab:relative_error_delta_on_perturbed}, we can better investigate effective defenses.
In Figure~\ref{fig:line_plots_defense_effectiveness}, we plot the error rates of each defense when paired with clean (\te) or a mixed combination of clean and poisoned (\teMixedDefinition) examples. 
While the ELMo \tr models in Table \ref{tab:relative_error_delta_on_perturbed} had the highest relative differences in error, these models outperform the same BERT and GloVe models.
The best defenses (\ie with the lowest error rates) are the same models that were most consistently generalizable across attack types -- \EnsChar and \EnsCharWord. 
\textit{These results indicate that defenses that include character-perturbed training data (\EnsChar and \EnsCharWord) are the most effective against character- and word-based attacks.}

\subsection{Attack Viewpoint}
Next, we examine susceptibility to adversarial attacks from the view of the attacker.   
We consider the impact on model confidence, and we analyze how a given attack impacts the uncertainty of classifications overall. For example, in a human-machine teaming scenario, a deception detection model would be used to flag content for a human fact-checker who may rely on the model's confidence when choosing whether to trust the classification.

In Figure \ref{fig:confidence_dists}, KDE plots illustrate model confidence distributions across examples from three test sets. We find that \teFullWord peaks at lower model confidences and flattens out as model confidence increases. 
By contrast, \te and \teFullChar peak at a model confidence close to 1. 
This shows that there is more confusion for predictions made on word-perturbed test data.  
If an analyst or end user relies on model confidence when choosing to accept a prediction, a significant difference in uncertainty of model classification can affect that decision. 
For example, when testing on clean examples (\te), the shift to a lower overall confidence may be enough to degrade the efficacy of the recommendation, even if the model has correctly classified the example.

\begin{figure}[t]
    \centering
\pgfplotstableread[col sep=comma]{figs/data/distribution_across_model_confidence_by_test.csv}\data  
 
\footnotesize
\begin{tikzpicture} 
\begin{axis}[  
    tick align=center,
    xtick pos=left,
    xlabel=Model Confidence,
    xlabel style={yshift=.5\baselineskip},
    ylabel style={yshift=-1.5\baselineskip},
    ylabel=KDE,
    height=1.2in,
    width=2.5in,
    xmin=0.1,xmax=1.1,
    ymin=0,ymax=4,
    scaled x ticks=false, 
    scaled y ticks=false, 
    every tick label/.append style={font=\tiny}, 
    legend style={fill=none, 
    at={(1.2,1)}, 
	anchor=north,legend columns=1,
	column sep=.1cm,draw=none}, 
	font = \footnotesize,
    ]

	 \addplot[BertErrors90, no marks, thick] table [x=modelConfidence,y=percentte]\data;
	 
	 \addplot[BertErrors90C, thick] table [x=modelConfidence,y=percentteC]\data;
	 
	 \addplot[BertErrors90W, thick] table [x=modelConfidence,y=percentteW]\data;
	   
\addlegendentry{\footnotesize \te};
\addlegendentry{\teFullChar};
\addlegendentry{\teFullWord};

\end{axis}
 \end{tikzpicture} 
 \vspace{-0.5\baselineskip}

    \caption{Kernel density estimation (KDE) plots illustrating distribution of model confidences.} 
    \label{fig:confidence_dists}
\end{figure}
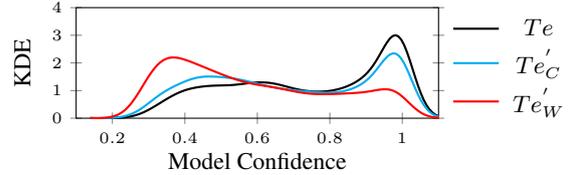

\begin{figure}[t!]
	\centering

\usepgfplotslibrary{colorbrewer} 
\pgfplotsset{compat = 1.15, cycle list/Set1-8}  
\usetikzlibrary{pgfplots.statistics, pgfplots.colorbrewer}  
\pgfplotsset{compat = 1.15}
\begin{tikzpicture}
	\pgfplotstableread[col sep=comma]{figs/tikz_data/fig1radar3way.csv}\csvdata
	\begin{axis}[
	title=\textit{\footnotesize 3-way Classification},
	title style={yshift=-\baselineskip},
		boxplot/draw direction = y,
		axis x line* = bottom,
		axis y line* = left,
		enlarge y limits,
		xticklabel style = {align=center, font=\tiny}, 
		yticklabel style = {align=center, font=\tiny}, 
		xtick = {1.5,3.5,5.5},
		xticklabels = {BERT,ELMo,GloVe},
		ylabel = {Error Rate},
		height=1.4in, width=1.75in,
		ymin=0.5,ymax=0.8,
	]
	
	\addplot+[boxplot, fill=radarTEc, draw=black] table[y=bertTEc] {\csvdata};
	\addplot+[boxplot, fill=radarTEw, draw=black] table[y=bertTEw] {\csvdata};
	\addplot+[boxplot, fill=radarTEc, draw=black] table[y=elmoTEc] {\csvdata};
	\addplot+[boxplot, fill=radarTEw, draw=black] table[y=elmoTEw] {\csvdata};
	\addplot+[boxplot, fill=radarTEc, draw=black] table[y=gloveTEc] {\csvdata};
	\addplot+[boxplot, fill=radarTEw, draw=black] table[y=gloveTEw] {\csvdata};
	
	\end{axis}
\end{tikzpicture}
\begin{tikzpicture}
	\pgfplotstableread[col sep=comma]{figs/tikz_data/fig1radar4way.csv}\csvdata
	\begin{axis}[
	title=\textit{\footnotesize 4-way Classification},
	title style={yshift=-\baselineskip},
		boxplot/draw direction = y,
		axis x line* = bottom,
		axis y line* = left,
		enlarge y limits,
		xticklabel style = {align=center, font=\tiny}, 
		yticklabel style = {align=center, font=\tiny}, 
		xtick = {1.5,3.5,5.5},
		xticklabels = {BERT,ELMo,GloVe},
		height=1.4in, width=1.75in,
		ymin=0.5,ymax=0.8,
	]
	\addplot+[boxplot, fill=radarTEc, draw=black] table[y=bertTEc] {\csvdata};
	\addplot+[boxplot, fill=radarTEw, draw=black] table[y=bertTEw] {\csvdata};
	\addplot+[boxplot, fill=radarTEc, draw=black] table[y=elmoTEc] {\csvdata};
	\addplot+[boxplot, fill=radarTEw, draw=black] table[y=elmoTEw] {\csvdata};
	\addplot+[boxplot, fill=radarTEc, draw=black] table[y=gloveTEc] {\csvdata};
	\addplot+[boxplot, fill=radarTEw, draw=black] table[y=gloveTEw] {\csvdata};
	
	\end{axis}
\end{tikzpicture}

\vspace{-0.5\baselineskip}
\footnotesize
 \begin{center}
 	\begin{tikzpicture}
 	\begin{axis}[
 	hide axis,
 	height = .75in,width=1in,
 	xmin = 0, xmax = 50, ymin = 0, ymax = 0.4,
 	legend cell align = {left}, legend columns=-1, legend style = {column sep=0.05in,
 	draw=none, at = {(0,0.75)}},]
 	
 	\addlegendimage{radarTEc, only marks, mark=square*, radarTEc} 
 	\addlegendentry{Character Attack~~ ($Te'_{C}$) ~~~~};
 	\addlegendimage{radarTEw, only marks, mark=square*, radarTEw}
 	\addlegendentry{Word Attack~~($Te'_{W}$ ) }; 
 	 
 	\end{axis}
 	\end{tikzpicture} 
 \end{center}

\vspace{-0.25\baselineskip}
 
	\vspace{-.75\baselineskip} 
    \vspace{-0.5\baselineskip}
	\caption{
	Box plots showing the effectiveness of the character  and word perturbation attack tactics via error rates across BERT-, ELMo-, and GloVe-based models.  
	}
	\label{fig:radar_plots} 
\end{figure}
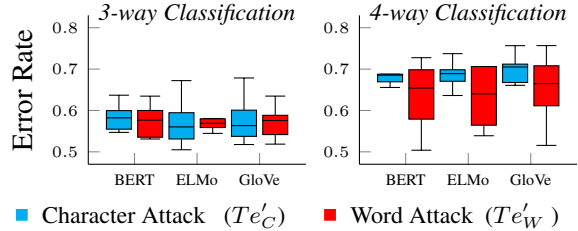

\begin{figure*}[t]
    \centering
    \includegraphics[scale=.43]{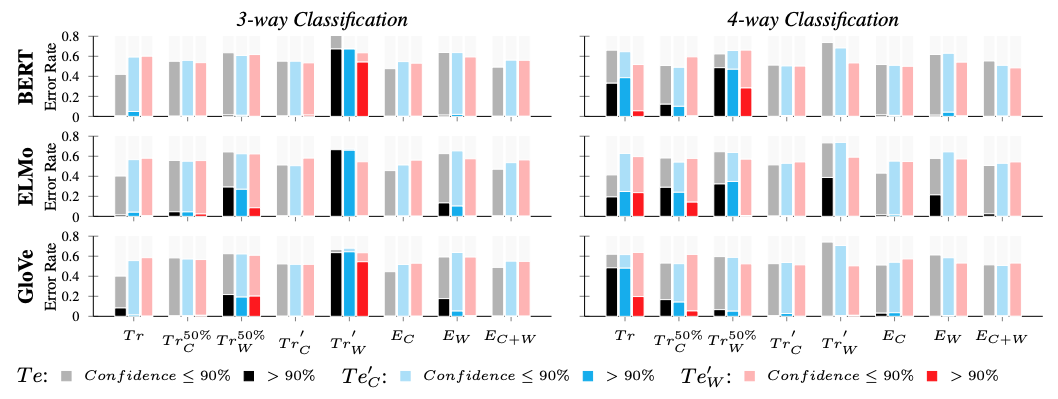}
     \caption{Error rates highlighting the prevalence of \textit{high confidence} ($>90\%$) errors when tested on each dataset. 
    }
    \label{fig:high_conf_bar}
\end{figure*}

Having examined the impact of attacks on model confidence, we next compare the effectiveness of each attack tactic when success is defined by the number of misclassifications. 
In Figure~\ref{fig:radar_plots}, box plots show the number of misclassifications as error rates.  
\teFullChar and \teFullWord attacks achieve similar median error rates in the 3-way task, and the maximum error rates are greater for the character than the word attacks. 
Although the 4-way task shows more discrepancies across attacks, again we see the character attacks display larger rates of error.
\textit{With both tasks, we see the largest number of misclassifications typically result from character-based attacks.}

Of note, the impact on the 3-way task is consistent across embedding types and attacks (the median error rates range from 56\% to 59\%). 
We see the widest range and largest maximum error rate with ELMo- and GloVe-based models when attacked with character-perturbed text.
Contrastly, the 4-way task displays similar trends across embeddings but not across attack types.
Although we see a greater range of error rates with the word attacks, the character attacks achieve a larger median and mean error rates than either \te or \teFullWord.

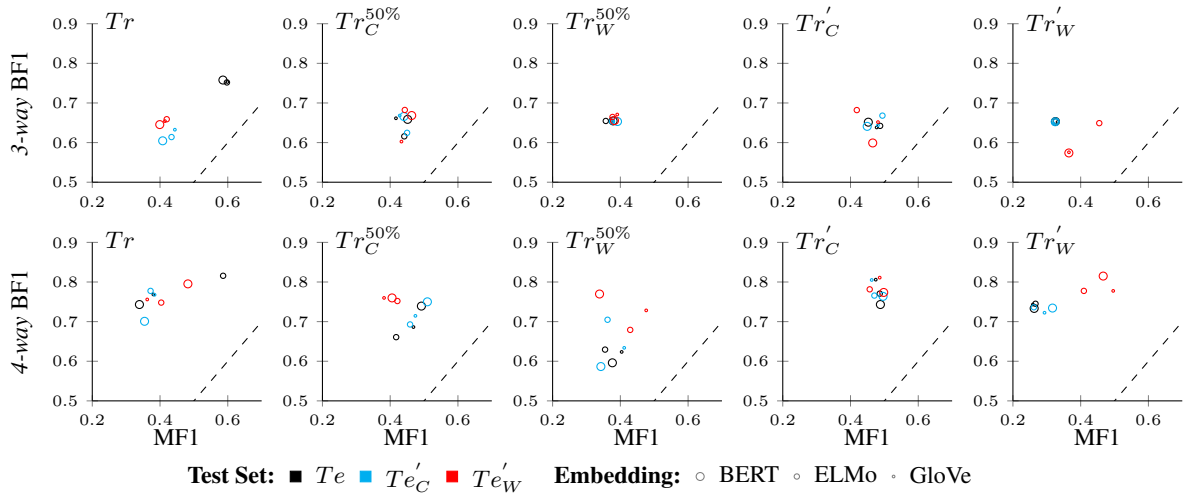
\begin{figure*}[th!]
    \centering 
    
\small 

\newcommand{\markSame}{o}
\newcommand{\bertMark}{\markSame}
\newcommand{\elmoMark}{\markSame}
\newcommand{\gloveMark}{\markSame}

\newcommand{\bertsize}{1.5pt}
\newcommand{\elmosize}{1pt}
\newcommand{\glovesize}{0.5pt}
\newcommand{\markopacity}{1}

\newcommand{\xmin}{0.2}
\newcommand{\xmax}{0.7}
\newcommand{\ymin}{0.5}
\newcommand{\ymax}{0.9}

\centering
\begin{tikzpicture}
\begin{groupplot}[
     group style = {
     group size = 5 by 1,
     horizontal sep=0.8cm 
     },
	height = 1.45in,
	width = 1.5in,
	xtick align = center, ytick align = center, 
	xticklabel style={font=\tiny},
	yticklabel style={font=\tiny},
	axis lines*=left, 
	xmin=\xmin,xmax=\xmax,ymin=\ymin, ymax=\ymax,
	xlabel style={font=\footnotesize},
	ylabel style={font=\footnotesize,yshift=-0.5\baselineskip}, 
	title style={yshift=-0.5\baselineskip,xshift=-3\baselineskip, anchor=west}
	]

    \newcommand{\multicolVal}{F13wayMicro}
    \newcommand{\tablePathVal}{figs/data/f1_results_3way.csv}
    \nextgroupplot[title={\bf \tr},  ylabel={ \textit{3-way} BF1}, ylabel style={align=center,yshift=-0.05in}]
    
            \addplot[black, dashed] coordinates {(0,0) (1,1)};
            	  
            \newcommand{\testVar}{clean}{  
            	 \addplot[ Errors90, mark=\bertMark, only marks, mark size=\bertsize,  ] table [col sep=comma,
            	 y=F1BinaryMicro, 
            	 x=\multicolVal,
            	 discard if not={Embedding}{bert}, 
            	 discard if not={Defense}{none}, 
            	 discard if not={Test}{\testVar}
            	 ]{\tablePathVal};
            	  
            	 \addplot[ Errors90, mark=\elmoMark, only marks, mark size=\elmosize,  ] table [col sep=comma,
            	 y=F1BinaryMicro, 
            	 x=\multicolVal, 
            	 discard if not={Embedding}{elmo}, 
            	 discard if not={Defense}{none}, 
            	 discard if not={Test}{\testVar}
            	 ]{\tablePathVal};
            	 
            	 \addplot[ Errors90, mark=\gloveMark, only marks, mark size=\glovesize,  ] table [col sep=comma,
            	 y=F1BinaryMicro, 
            	 x=\multicolVal, 
            	 discard if not={Embedding}{glove}, 
            	 discard if not={Defense}{none}, 
            	 discard if not={Test}{\testVar}
            	 ]{\tablePathVal};
            }
            
            \renewcommand{\testVar}{100char}{ 
            	 \addplot[CErrors90, mark=\bertMark, only marks, mark size=\bertsize,  ] table [col sep=comma,
            	 y=F1BinaryMicro, 
            	 x=\multicolVal,
            	 discard if not={Embedding}{bert}, 
            	 discard if not={Defense}{none}, 
            	 discard if not={Test}{\testVar}
            	 ]{\tablePathVal};
            	  
            	 \addplot[CErrors90, mark=\elmoMark, only marks, mark size=\elmosize,  ] table [col sep=comma,
            	 y=F1BinaryMicro, 
            	 x=\multicolVal, 
            	 discard if not={Embedding}{elmo}, 
            	 discard if not={Defense}{none}, 
            	 discard if not={Test}{\testVar}
            	 ]{\tablePathVal};
            	 
            	 \addplot[CErrors90, mark=\gloveMark, only marks, mark size=\glovesize,  ] table [col sep=comma,
            	 y=F1BinaryMicro, 
            	 x=\multicolVal, 
            	 discard if not={Embedding}{glove}, 
            	 discard if not={Defense}{none}, 
            	 discard if not={Test}{\testVar}
            	 ]{\tablePathVal};
            }
                	 
            \renewcommand{\testVar}{100word}{ 
            	 \addplot[WErrors90, mark=\bertMark, only marks, mark size=\bertsize,  ] table [col sep=comma,
            	 y=F1BinaryMicro, 
            	 x=\multicolVal,
            	 discard if not={Embedding}{bert}, 
            	 discard if not={Defense}{none}, 
            	 discard if not={Test}{\testVar}
            	 ]{\tablePathVal};
            	  
            	 \addplot[WErrors90, mark=\elmoMark, only marks, mark size=\elmosize,  ] table [col sep=comma,
            	 y=F1BinaryMicro, 
            	 x=\multicolVal, 
            	 discard if not={Embedding}{elmo}, 
            	 discard if not={Defense}{none}, 
            	 discard if not={Test}{\testVar}
            	 ]{\tablePathVal};
            	 
            	 \addplot[WErrors90, mark=\gloveMark, only marks, mark size=\glovesize,  ] table [col sep=comma,
            	 y=F1BinaryMicro, 
            	 x=\multicolVal, 
            	 discard if not={Embedding}{glove}, 
            	 discard if not={Defense}{none}, 
            	 discard if not={Test}{\testVar}
            	 ]{\tablePathVal};
            }

    \nextgroupplot[title={{\bf \trHalfChar}}]
            \addplot[black, dashed] coordinates {(0,0) (1,1)};
            
            \newcommand{\testVar}{clean} 
            	 \addplot[ Errors90, mark=\bertMark, only marks, mark size=\bertsize,  ] table [col sep=comma,
            	 y=F1BinaryMicro, 
            	 x=\multicolVal,
            	 discard if not={Embedding}{bert}, 
            	 discard if not={Defense}{50char}, 
            	 discard if not={Test}{\testVar}
            	 ]{\tablePathVal};
            	  
            	 \addplot[ Errors90, mark=\elmoMark, only marks, mark size=\elmosize,  ] table [col sep=comma,
            	 y=F1BinaryMicro, 
            	 x=\multicolVal, 
            	 discard if not={Embedding}{elmo}, 
            	 discard if not={Defense}{50char}, 
            	 discard if not={Test}{\testVar}
            	 ]{\tablePathVal};
            	 
            	 \addplot[ Errors90, mark=\gloveMark, only marks, mark size=\glovesize,  ] table [col sep=comma,
            	 y=F1BinaryMicro, 
            	 x=\multicolVal, 
            	 discard if not={Embedding}{glove}, 
            	 discard if not={Defense}{50char}, 
            	 discard if not={Test}{\testVar}
            	 ]{\tablePathVal};
            
            \renewcommand{\testVar}{100char} 
            	 \addplot[CErrors90, mark=\bertMark, only marks, mark size=\bertsize,  ] table [col sep=comma,
            	 y=F1BinaryMicro, 
            	 x=\multicolVal,
            	 discard if not={Embedding}{bert}, 
            	 discard if not={Defense}{50char}, 
            	 discard if not={Test}{\testVar}
            	 ]{\tablePathVal};
            	  
            	 \addplot[CErrors90, mark=\elmoMark, only marks, mark size=\elmosize,  ] table [col sep=comma,
            	 y=F1BinaryMicro, 
            	 x=\multicolVal, 
            	 discard if not={Embedding}{elmo}, 
            	 discard if not={Defense}{50char}, 
            	 discard if not={Test}{\testVar}
            	 ]{\tablePathVal};
            	 
            	 \addplot[CErrors90, mark=\gloveMark, only marks, mark size=\glovesize,  ] table [col sep=comma,
            	 y=F1BinaryMicro, 
            	 x=\multicolVal, 
            	 discard if not={Embedding}{glove}, 
            	 discard if not={Defense}{50char}, 
            	 discard if not={Test}{\testVar}
            	 ]{\tablePathVal};
                	 
            \renewcommand{\testVar}{100word} 
            	 \addplot[WErrors90, mark=\bertMark, only marks, mark size=\bertsize,  ] table [col sep=comma,
            	 y=F1BinaryMicro, 
            	 x=\multicolVal,
            	 discard if not={Embedding}{bert}, 
            	 discard if not={Defense}{50char}, 
            	 discard if not={Test}{\testVar}
            	 ]{\tablePathVal};
            	  
            	 \addplot[WErrors90, mark=\elmoMark, only marks, mark size=\elmosize,  ] table [col sep=comma,
            	 y=F1BinaryMicro, 
            	 x=\multicolVal, 
            	 discard if not={Embedding}{elmo}, 
            	 discard if not={Defense}{50char}, 
            	 discard if not={Test}{\testVar}
            	 ]{\tablePathVal};
            	 
            	 \addplot[WErrors90, mark=\gloveMark, only marks, mark size=\glovesize,  ] table [col sep=comma,
            	 y=F1BinaryMicro, 
            	 x=\multicolVal, 
            	 discard if not={Embedding}{glove}, 
            	 discard if not={Defense}{50char}, 
            	 discard if not={Test}{\testVar}
            	 ]{\tablePathVal};

            \nextgroupplot[title={{\bf \trHalfWord}}]
            
                \addplot[black, dashed] coordinates {(0,0) (1,1)};
                
                    \newcommand{\testVar}{clean} 
                    	 \addplot[ Errors90, mark=\bertMark, only marks, mark size=\bertsize,  ] table [col sep=comma,
                    	 y=F1BinaryMicro, 
                    	 x=\multicolVal,
                    	 discard if not={Embedding}{bert}, 
                    	 discard if not={Defense}{50word}, 
                    	 discard if not={Test}{\testVar}
                    	 ]{\tablePathVal};
                    	  
                    	 \addplot[ Errors90, mark=\elmoMark, only marks, mark size=\elmosize,  ] table [col sep=comma,
                    	 y=F1BinaryMicro, 
                    	 x=\multicolVal, 
                    	 discard if not={Embedding}{elmo}, 
                    	 discard if not={Defense}{50word}, 
                    	 discard if not={Test}{\testVar}
                    	 ]{\tablePathVal};
                    	 
                    	 \addplot[ Errors90, mark=\gloveMark, only marks, mark size=\glovesize,  ] table [col sep=comma,
                    	 y=F1BinaryMicro, 
                    	 x=\multicolVal, 
                    	 discard if not={Embedding}{glove}, 
                    	 discard if not={Defense}{50word}, 
                    	 discard if not={Test}{\testVar}
                    	 ]{\tablePathVal};
                    
                    \renewcommand{\testVar}{100char} 
                    	 \addplot[CErrors90, mark=\bertMark, only marks, mark size=\bertsize,  ] table [col sep=comma,
                    	 y=F1BinaryMicro, 
                    	 x=\multicolVal,
                    	 discard if not={Embedding}{bert}, 
                    	 discard if not={Defense}{50word}, 
                    	 discard if not={Test}{\testVar}
                    	 ]{\tablePathVal};
                    	  
                    	 \addplot[CErrors90, mark=\elmoMark, only marks, mark size=\elmosize,  ] table [col sep=comma,
                    	 y=F1BinaryMicro, 
                    	 x=\multicolVal, 
                    	 discard if not={Embedding}{elmo}, 
                    	 discard if not={Defense}{50word}, 
                    	 discard if not={Test}{\testVar}
                    	 ]{\tablePathVal};
                    	 
                    	 \addplot[CErrors90, mark=\gloveMark, only marks, mark size=\glovesize,  ] table [col sep=comma,
                    	 y=F1BinaryMicro, 
                    	 x=\multicolVal, 
                    	 discard if not={Embedding}{glove}, 
                    	 discard if not={Defense}{50word}, 
                    	 discard if not={Test}{\testVar}
                    	 ]{\tablePathVal};
                        	 
                    \renewcommand{\testVar}{100word} 
                    	 \addplot[WErrors90, mark=\bertMark, only marks, mark size=\bertsize,  ] table [col sep=comma,
                    	 y=F1BinaryMicro, 
                    	 x=\multicolVal,
                    	 discard if not={Embedding}{bert}, 
                    	 discard if not={Defense}{50word}, 
                    	 discard if not={Test}{\testVar}
                    	 ]{\tablePathVal};
                    	  
                    	 \addplot[WErrors90, mark=\elmoMark, only marks, mark size=\elmosize,  ] table [col sep=comma,
                    	 y=F1BinaryMicro, 
                    	 x=\multicolVal, 
                    	 discard if not={Embedding}{elmo}, 
                    	 discard if not={Defense}{50word}, 
                    	 discard if not={Test}{\testVar}
                    	 ]{\tablePathVal};
                    	 
                    	 \addplot[WErrors90, mark=\gloveMark, only marks, mark size=\glovesize,  ] table [col sep=comma,
                    	 y=F1BinaryMicro, 
                    	 x=\multicolVal, 
                    	 discard if not={Embedding}{glove}, 
                    	 discard if not={Defense}{50word}, 
                    	 discard if not={Test}{\testVar}
                    	 ]{\tablePathVal};

    \nextgroupplot[title={{\bf \trFullChar}}]
    
                \addplot[black, dashed] coordinates {(0,0) (1,1)};

            \newcommand{\testVar}{clean} 
            	 \addplot[ Errors90, mark=\bertMark, only marks, mark size=\bertsize,  ] table [col sep=comma,
            	 y=F1BinaryMicro, 
            	 x=\multicolVal,
            	 discard if not={Embedding}{bert}, 
            	 discard if not={Defense}{100char}, 
            	 discard if not={Test}{\testVar}
            	 ]{\tablePathVal};
            	  
            	 \addplot[ Errors90, mark=\elmoMark, only marks, mark size=\elmosize,  ] table [col sep=comma,
            	 y=F1BinaryMicro, 
            	 x=\multicolVal, 
            	 discard if not={Embedding}{elmo}, 
            	 discard if not={Defense}{100char}, 
            	 discard if not={Test}{\testVar}
            	 ]{\tablePathVal};
            	 
            	 \addplot[ Errors90, mark=\gloveMark, only marks, mark size=\glovesize,  ] table [col sep=comma,
            	 y=F1BinaryMicro, 
            	 x=\multicolVal, 
            	 discard if not={Embedding}{glove}, 
            	 discard if not={Defense}{100char}, 
            	 discard if not={Test}{\testVar}
            	 ]{\tablePathVal};
            
            \renewcommand{\testVar}{100char} 
            	 \addplot[CErrors90, mark=\bertMark, only marks, mark size=\bertsize,  ] table [col sep=comma,
            	 y=F1BinaryMicro, 
            	 x=\multicolVal,
            	 discard if not={Embedding}{bert}, 
            	 discard if not={Defense}{100char}, 
            	 discard if not={Test}{\testVar}
            	 ]{\tablePathVal};
            	  
            	 \addplot[CErrors90, mark=\elmoMark, only marks, mark size=\elmosize,  ] table [col sep=comma,
            	 y=F1BinaryMicro, 
            	 x=\multicolVal, 
            	 discard if not={Embedding}{elmo}, 
            	 discard if not={Defense}{100char}, 
            	 discard if not={Test}{\testVar}
            	 ]{\tablePathVal};
            	 
            	 \addplot[CErrors90, mark=\gloveMark, only marks, mark size=\glovesize,  ] table [col sep=comma,
            	 y=F1BinaryMicro, 
            	 x=\multicolVal, 
            	 discard if not={Embedding}{glove}, 
            	 discard if not={Defense}{100char}, 
            	 discard if not={Test}{\testVar}
            	 ]{\tablePathVal};
                	 
            \renewcommand{\testVar}{100word} 
            	 \addplot[WErrors90, mark=\bertMark, only marks, mark size=\bertsize,  ] table [col sep=comma,
            	 y=F1BinaryMicro, 
            	 x=\multicolVal,
            	 discard if not={Embedding}{bert}, 
            	 discard if not={Defense}{100char}, 
            	 discard if not={Test}{\testVar}
            	 ]{\tablePathVal};
            	  
            	 \addplot[WErrors90, mark=\elmoMark, only marks, mark size=\elmosize,  ] table [col sep=comma,
            	 y=F1BinaryMicro, 
            	 x=\multicolVal, 
            	 discard if not={Embedding}{elmo}, 
            	 discard if not={Defense}{100char}, 
            	 discard if not={Test}{\testVar}
            	 ]{\tablePathVal};
            	 
            	 \addplot[WErrors90, mark=\gloveMark, only marks, mark size=\glovesize,  ] table [col sep=comma,
            	 y=F1BinaryMicro, 
            	 x=\multicolVal, 
            	 discard if not={Embedding}{glove}, 
            	 discard if not={Defense}{100char}, 
            	 discard if not={Test}{\testVar}
            	 ]{\tablePathVal};

    \nextgroupplot[title={{\bf \trFullWord}}]
    
                \addplot[black, dashed] coordinates {(0,0) (1,1)};

            \newcommand{\testVar}{clean} 
            	 \addplot[ Errors90, mark=\bertMark, only marks, mark size=\bertsize,  ] table [col sep=comma,
            	 y=F1BinaryMicro, 
            	 x=\multicolVal,
            	 discard if not={Embedding}{bert}, 
            	 discard if not={Defense}{100word}, 
            	 discard if not={Test}{\testVar}
            	 ]{\tablePathVal};
            	  
            	 \addplot[ Errors90, mark=\elmoMark, only marks, mark size=\elmosize,  ] table [col sep=comma,
            	 y=F1BinaryMicro, 
            	 x=\multicolVal, 
            	 discard if not={Embedding}{elmo}, 
            	 discard if not={Defense}{100word}, 
            	 discard if not={Test}{\testVar}
            	 ]{\tablePathVal};
            	 
            	 \addplot[ Errors90, mark=\gloveMark, only marks, mark size=\glovesize,  ] table [col sep=comma,
            	 y=F1BinaryMicro, 
            	 x=\multicolVal, 
            	 discard if not={Embedding}{glove}, 
            	 discard if not={Defense}{100word}, 
            	 discard if not={Test}{\testVar}
            	 ]{\tablePathVal};
          
         { 
            \renewcommand{\testVar}{100char} 
            	 \addplot[CErrors90, mark=\bertMark, only marks, mark size=\bertsize,  ] table [col sep=comma,
            	 y=F1BinaryMicro, 
            	 x=\multicolVal,
            	 discard if not={Embedding}{bert}, 
            	 discard if not={Defense}{100word}, 
            	 discard if not={Test}{\testVar}
            	 ]{\tablePathVal};
            	  
            	 \addplot[CErrors90, mark=\elmoMark, only marks, mark size=\elmosize,  ] table [col sep=comma,
            	 y=F1BinaryMicro, 
            	 x=\multicolVal, 
            	 discard if not={Embedding}{elmo}, 
            	 discard if not={Defense}{100word}, 
            	 discard if not={Test}{\testVar}
            	 ]{\tablePathVal};
            	 
            	 \addplot[CErrors90, mark=\gloveMark, only marks, mark size=\glovesize,  ] table [col sep=comma,
            	 y=F1BinaryMicro, 
            	 x=\multicolVal, 
            	 discard if not={Embedding}{glove}, 
            	 discard if not={Defense}{100word}, 
            	 discard if not={Test}{\testVar}
            	 ]{\tablePathVal};
                	 
            \renewcommand{\testVar}{100word} 
            	 \addplot[WErrors90, mark=\bertMark, only marks, mark size=\bertsize,  ] table [col sep=comma,
            	 y=F1BinaryMicro, 
            	 x=\multicolVal,
            	 discard if not={Embedding}{bert}, 
            	 discard if not={Defense}{100word}, 
            	 discard if not={Test}{\testVar}
            	 ]{\tablePathVal};
            	  
            	 \addplot[WErrors90, mark=\elmoMark, only marks, mark size=\elmosize,  ] table [col sep=comma,
            	 y=F1BinaryMicro, 
            	 x=\multicolVal, 
            	 discard if not={Embedding}{elmo}, 
            	 discard if not={Defense}{100word}, 
            	 discard if not={Test}{\testVar}
            	 ]{\tablePathVal};
            	 
            	 \addplot[WErrors90, mark=\gloveMark, only marks, mark size=\glovesize,  ] table [col sep=comma,
            	 y=F1BinaryMicro, 
            	 x=\multicolVal, 
            	 discard if not={Embedding}{glove}, 
            	 discard if not={Defense}{100word}, 
            	 discard if not={Test}{\testVar}
            	 ]{\tablePathVal};

        }

\end{groupplot}
\end{tikzpicture} 
\vspace{-0.4\baselineskip}
\begin{tikzpicture}
\begin{groupplot}[
     group style = {
     group size = 5 by 1,
     horizontal sep=0.8cm 
     }, 
	height = 1.45in,
	width = 1.5in,
	xtick align = center, ytick align = center, 
	xticklabel style={font=\tiny},
	yticklabel style={font=\tiny},
	axis lines*=left, 
	xmin=\xmin,xmax=\xmax,ymin=\ymin, ymax=\ymax,
	xlabel style={font=\footnotesize,yshift=0.75\baselineskip},
	ylabel style={font=\footnotesize, yshift=-0.5\baselineskip}, 
	title style={yshift=-0.5\baselineskip,xshift=-3\baselineskip, anchor=west},
	xlabel=MF1,
	]

    \newcommand{\multicolVal}{F14wayMicro}
    \newcommand{\tablePathVal}{figs/data/f1_results_4way.csv}
    \nextgroupplot[ylabel={\textit{4-way} BF1}, ylabel style={align=center,yshift=-0.05in}, title=\tr]
    
                \addplot[black, dashed] coordinates {(0,0) (1,1)};

            \newcommand{\testVar}{clean}{
            	 \addplot[ Errors90, mark=\bertMark, only marks, mark size=\bertsize,  ] table [col sep=comma,
            	 y=F1BinaryMicro, 
            	 x=\multicolVal,
            	 discard if not={Embedding}{bert}, 
            	 discard if not={Defense}{none}, 
            	 discard if not={Test}{\testVar}
            	 ]{\tablePathVal};
            	  
            	 \addplot[ Errors90, mark=\elmoMark, only marks, mark size=\elmosize,  ] table [col sep=comma,
            	 y=F1BinaryMicro, 
            	 x=\multicolVal, 
            	 discard if not={Embedding}{elmo}, 
            	 discard if not={Defense}{none}, 
            	 discard if not={Test}{\testVar}
            	 ]{\tablePathVal};
            	 
            	 \addplot[ Errors90, mark=\gloveMark, only marks, mark size=\glovesize,  ] table [col sep=comma,
            	 y=F1BinaryMicro, 
            	 x=\multicolVal, 
            	 discard if not={Embedding}{glove}, 
            	 discard if not={Defense}{none}, 
            	 discard if not={Test}{\testVar}
            	 ]{\tablePathVal};
            }
            
            \renewcommand{\testVar}{100char}{ 
            	 \addplot[CErrors90, mark=\bertMark, only marks, mark size=\bertsize,  ] table [col sep=comma,
            	 y=F1BinaryMicro, 
            	 x=\multicolVal,
            	 discard if not={Embedding}{bert}, 
            	 discard if not={Defense}{none}, 
            	 discard if not={Test}{\testVar}
            	 ]{\tablePathVal};
            	  
            	 \addplot[CErrors90, mark=\elmoMark, only marks, mark size=\elmosize,  ] table [col sep=comma,
            	 y=F1BinaryMicro, 
            	 x=\multicolVal, 
            	 discard if not={Embedding}{elmo}, 
            	 discard if not={Defense}{none}, 
            	 discard if not={Test}{\testVar}
            	 ]{\tablePathVal};
            	 
            	 \addplot[CErrors90, mark=\gloveMark, only marks, mark size=\glovesize,  ] table [col sep=comma,
            	 y=F1BinaryMicro, 
            	 x=\multicolVal, 
            	 discard if not={Embedding}{glove}, 
            	 discard if not={Defense}{none}, 
            	 discard if not={Test}{\testVar}
            	 ]{\tablePathVal};
            }
                	 
            \renewcommand{\testVar}{100word}{ 
            	 \addplot[WErrors90, mark=\bertMark, only marks, mark size=\bertsize,  ] table [col sep=comma,
            	 y=F1BinaryMicro, 
            	 x=\multicolVal,
            	 discard if not={Embedding}{bert}, 
            	 discard if not={Defense}{none}, 
            	 discard if not={Test}{\testVar}
            	 ]{\tablePathVal};
            	  
            	 \addplot[WErrors90, mark=\elmoMark, only marks, mark size=\elmosize,  ] table [col sep=comma,
            	 y=F1BinaryMicro, 
            	 x=\multicolVal, 
            	 discard if not={Embedding}{elmo}, 
            	 discard if not={Defense}{none}, 
            	 discard if not={Test}{\testVar}
            	 ]{\tablePathVal};
            	 
            	 \addplot[WErrors90, mark=\gloveMark, only marks, mark size=\glovesize,  ] table [col sep=comma,
            	 y=F1BinaryMicro, 
            	 x=\multicolVal, 
            	 discard if not={Embedding}{glove}, 
            	 discard if not={Defense}{none}, 
            	 discard if not={Test}{\testVar}
            	 ]{\tablePathVal};
            }

    \nextgroupplot[title=\trHalfChar]
    
                \addplot[black, dashed] coordinates {(0,0) (1,1)};

            \newcommand{\testVar}{clean} 
            	 \addplot[ Errors90, mark=\bertMark, only marks, mark size=\bertsize,  ] table [col sep=comma,
            	 y=F1BinaryMicro, 
            	 x=\multicolVal,
            	 discard if not={Embedding}{bert}, 
            	 discard if not={Defense}{50char}, 
            	 discard if not={Test}{\testVar}
            	 ]{\tablePathVal};
            	  
            	 \addplot[ Errors90, mark=\elmoMark, only marks, mark size=\elmosize,  ] table [col sep=comma,
            	 y=F1BinaryMicro, 
            	 x=\multicolVal, 
            	 discard if not={Embedding}{elmo}, 
            	 discard if not={Defense}{50char}, 
            	 discard if not={Test}{\testVar}
            	 ]{\tablePathVal};
            	 
            	 \addplot[ Errors90, mark=\gloveMark, only marks, mark size=\glovesize,  ] table [col sep=comma,
            	 y=F1BinaryMicro, 
            	 x=\multicolVal, 
            	 discard if not={Embedding}{glove}, 
            	 discard if not={Defense}{50char}, 
            	 discard if not={Test}{\testVar}
            	 ]{\tablePathVal};
            
            \renewcommand{\testVar}{100char} 
            	 \addplot[CErrors90, mark=\bertMark, only marks, mark size=\bertsize,  ] table [col sep=comma,
            	 y=F1BinaryMicro, 
            	 x=\multicolVal,
            	 discard if not={Embedding}{bert}, 
            	 discard if not={Defense}{50char}, 
            	 discard if not={Test}{\testVar}
            	 ]{\tablePathVal};
            	  
            	 \addplot[CErrors90, mark=\elmoMark, only marks, mark size=\elmosize,  ] table [col sep=comma,
            	 y=F1BinaryMicro, 
            	 x=\multicolVal, 
            	 discard if not={Embedding}{elmo}, 
            	 discard if not={Defense}{50char}, 
            	 discard if not={Test}{\testVar}
            	 ]{\tablePathVal};
            	 
            	 \addplot[CErrors90, mark=\gloveMark, only marks, mark size=\glovesize,  ] table [col sep=comma,
            	 y=F1BinaryMicro, 
            	 x=\multicolVal, 
            	 discard if not={Embedding}{glove}, 
            	 discard if not={Defense}{50char}, 
            	 discard if not={Test}{\testVar}
            	 ]{\tablePathVal};
                	 
            \renewcommand{\testVar}{100word} 
            	 \addplot[WErrors90, mark=\bertMark, only marks, mark size=\bertsize,  ] table [col sep=comma,
            	 y=F1BinaryMicro, 
            	 x=\multicolVal,
            	 discard if not={Embedding}{bert}, 
            	 discard if not={Defense}{50char}, 
            	 discard if not={Test}{\testVar}
            	 ]{\tablePathVal};
            	  
            	 \addplot[WErrors90, mark=\elmoMark, only marks, mark size=\elmosize,  ] table [col sep=comma,
            	 y=F1BinaryMicro, 
            	 x=\multicolVal, 
            	 discard if not={Embedding}{elmo}, 
            	 discard if not={Defense}{50char}, 
            	 discard if not={Test}{\testVar}
            	 ]{\tablePathVal};
            	 
            	 \addplot[WErrors90, mark=\gloveMark, only marks, mark size=\glovesize,  ] table [col sep=comma,
            	 y=F1BinaryMicro, 
            	 x=\multicolVal, 
            	 discard if not={Embedding}{glove}, 
            	 discard if not={Defense}{50char}, 
            	 discard if not={Test}{\testVar}
            	 ]{\tablePathVal};

        \nextgroupplot[title=\trHalfWord]
        
                \addplot[black, dashed] coordinates {(0,0) (1,1)};

                    \newcommand{\testVar}{clean} 
                    	 \addplot[ Errors90, mark=\bertMark, only marks, mark size=\bertsize,  ] table [col sep=comma,
                    	 y=F1BinaryMicro, 
                    	 x=\multicolVal,
                    	 discard if not={Embedding}{bert}, 
                    	 discard if not={Defense}{50word}, 
                    	 discard if not={Test}{\testVar}
                    	 ]{\tablePathVal};
                    	  
                    	 \addplot[ Errors90, mark=\elmoMark, only marks, mark size=\elmosize,  ] table [col sep=comma,
                    	 y=F1BinaryMicro, 
                    	 x=\multicolVal, 
                    	 discard if not={Embedding}{elmo}, 
                    	 discard if not={Defense}{50word}, 
                    	 discard if not={Test}{\testVar}
                    	 ]{\tablePathVal};
                    	 
                    	 \addplot[ Errors90, mark=\gloveMark, only marks, mark size=\glovesize,  ] table [col sep=comma,
                    	 y=F1BinaryMicro, 
                    	 x=\multicolVal, 
                    	 discard if not={Embedding}{glove}, 
                    	 discard if not={Defense}{50word}, 
                    	 discard if not={Test}{\testVar}
                    	 ]{\tablePathVal};
                    
                    \renewcommand{\testVar}{100char} 
                    	 \addplot[CErrors90, mark=\bertMark, only marks, mark size=\bertsize,  ] table [col sep=comma,
                    	 y=F1BinaryMicro, 
                    	 x=\multicolVal,
                    	 discard if not={Embedding}{bert}, 
                    	 discard if not={Defense}{50word}, 
                    	 discard if not={Test}{\testVar}
                    	 ]{\tablePathVal};
                    	  
                    	 \addplot[CErrors90, mark=\elmoMark, only marks, mark size=\elmosize,  ] table [col sep=comma,
                    	 y=F1BinaryMicro, 
                    	 x=\multicolVal, 
                    	 discard if not={Embedding}{elmo}, 
                    	 discard if not={Defense}{50word}, 
                    	 discard if not={Test}{\testVar}
                    	 ]{\tablePathVal};
                    	 
                    	 \addplot[CErrors90, mark=\gloveMark, only marks, mark size=\glovesize,  ] table [col sep=comma,
                    	 y=F1BinaryMicro, 
                    	 x=\multicolVal, 
                    	 discard if not={Embedding}{glove}, 
                    	 discard if not={Defense}{50word}, 
                    	 discard if not={Test}{\testVar}
                    	 ]{\tablePathVal};
                        	 
                    \renewcommand{\testVar}{100word} 
                    	 \addplot[WErrors90, mark=\bertMark, only marks, mark size=\bertsize,  ] table [col sep=comma,
                    	 y=F1BinaryMicro, 
                    	 x=\multicolVal,
                    	 discard if not={Embedding}{bert}, 
                    	 discard if not={Defense}{50word}, 
                    	 discard if not={Test}{\testVar}
                    	 ]{\tablePathVal};
                    	  
                    	 \addplot[WErrors90, mark=\elmoMark, only marks, mark size=\elmosize,  ] table [col sep=comma,
                    	 y=F1BinaryMicro, 
                    	 x=\multicolVal, 
                    	 discard if not={Embedding}{elmo}, 
                    	 discard if not={Defense}{50word}, 
                    	 discard if not={Test}{\testVar}
                    	 ]{\tablePathVal};
                    	 
                    	 \addplot[WErrors90, mark=\gloveMark, only marks, mark size=\glovesize,  ] table [col sep=comma,
                    	 y=F1BinaryMicro, 
                    	 x=\multicolVal, 
                    	 discard if not={Embedding}{glove}, 
                    	 discard if not={Defense}{50word}, 
                    	 discard if not={Test}{\testVar}
                    	 ]{\tablePathVal};

    \nextgroupplot[title=\trFullChar]
    
                \addplot[black, dashed] coordinates {(0,0) (1,1)};

            \newcommand{\testVar}{clean} 
            	 \addplot[ Errors90, mark=\bertMark, only marks, mark size=\bertsize,  ] table [col sep=comma,
            	 y=F1BinaryMicro, 
            	 x=\multicolVal,
            	 discard if not={Embedding}{bert}, 
            	 discard if not={Defense}{100char}, 
            	 discard if not={Test}{\testVar}
            	 ]{\tablePathVal};
            	  
            	 \addplot[ Errors90, mark=\elmoMark, only marks, mark size=\elmosize,  ] table [col sep=comma,
            	 y=F1BinaryMicro, 
            	 x=\multicolVal, 
            	 discard if not={Embedding}{elmo}, 
            	 discard if not={Defense}{100char}, 
            	 discard if not={Test}{\testVar}
            	 ]{\tablePathVal};
            	 
            	 \addplot[ Errors90, mark=\gloveMark, only marks, mark size=\glovesize,  ] table [col sep=comma,
            	 y=F1BinaryMicro, 
            	 x=\multicolVal, 
            	 discard if not={Embedding}{glove}, 
            	 discard if not={Defense}{100char}, 
            	 discard if not={Test}{\testVar}
            	 ]{\tablePathVal};
            
            \renewcommand{\testVar}{100char} 
            	 \addplot[CErrors90, mark=\bertMark, only marks, mark size=\bertsize,  ] table [col sep=comma,
            	 y=F1BinaryMicro, 
            	 x=\multicolVal,
            	 discard if not={Embedding}{bert}, 
            	 discard if not={Defense}{100char}, 
            	 discard if not={Test}{\testVar}
            	 ]{\tablePathVal};
            	  
            	 \addplot[CErrors90, mark=\elmoMark, only marks, mark size=\elmosize,  ] table [col sep=comma,
            	 y=F1BinaryMicro, 
            	 x=\multicolVal, 
            	 discard if not={Embedding}{elmo}, 
            	 discard if not={Defense}{100char}, 
            	 discard if not={Test}{\testVar}
            	 ]{\tablePathVal};
            	 
            	 \addplot[CErrors90, mark=\gloveMark, only marks, mark size=\glovesize,  ] table [col sep=comma,
            	 y=F1BinaryMicro, 
            	 x=\multicolVal, 
            	 discard if not={Embedding}{glove}, 
            	 discard if not={Defense}{100char}, 
            	 discard if not={Test}{\testVar}
            	 ]{\tablePathVal};
                	 
            \renewcommand{\testVar}{100word} 
            	 \addplot[WErrors90, mark=\bertMark, only marks, mark size=\bertsize,  ] table [col sep=comma,
            	 y=F1BinaryMicro, 
            	 x=\multicolVal,
            	 discard if not={Embedding}{bert}, 
            	 discard if not={Defense}{100char}, 
            	 discard if not={Test}{\testVar}
            	 ]{\tablePathVal};
            	  
            	 \addplot[WErrors90, mark=\elmoMark, only marks, mark size=\elmosize,  ] table [col sep=comma,
            	 y=F1BinaryMicro, 
            	 x=\multicolVal, 
            	 discard if not={Embedding}{elmo}, 
            	 discard if not={Defense}{100char}, 
            	 discard if not={Test}{\testVar}
            	 ]{\tablePathVal};
            	 
            	 \addplot[WErrors90, mark=\gloveMark, only marks, mark size=\glovesize,  ] table [col sep=comma,
            	 y=F1BinaryMicro, 
            	 x=\multicolVal, 
            	 discard if not={Embedding}{glove}, 
            	 discard if not={Defense}{100char}, 
            	 discard if not={Test}{\testVar}
            	 ]{\tablePathVal};

    \nextgroupplot[title=\trFullWord]
    
                \addplot[black, dashed] coordinates {(0,0) (1,1)};

            \newcommand{\testVar}{clean} 
            	 \addplot[ Errors90, mark=\bertMark, only marks, mark size=\bertsize,  ] table [col sep=comma,
            	 y=F1BinaryMicro, 
            	 x=\multicolVal,
            	 discard if not={Embedding}{bert}, 
            	 discard if not={Defense}{100word}, 
            	 discard if not={Test}{\testVar}
            	 ]{\tablePathVal};
            	  
            	 \addplot[ Errors90, mark=\elmoMark, only marks, mark size=\elmosize,  ] table [col sep=comma,
            	 y=F1BinaryMicro, 
            	 x=\multicolVal, 
            	 discard if not={Embedding}{elmo}, 
            	 discard if not={Defense}{100word}, 
            	 discard if not={Test}{\testVar}
            	 ]{\tablePathVal};
            	 
            	 \addplot[ Errors90, mark=\gloveMark, only marks, mark size=\glovesize,  ] table [col sep=comma,
            	 y=F1BinaryMicro, 
            	 x=\multicolVal, 
            	 discard if not={Embedding}{glove}, 
            	 discard if not={Defense}{100word}, 
            	 discard if not={Test}{\testVar}
            	 ]{\tablePathVal};
            
            \renewcommand{\testVar}{100char} 
            	 \addplot[CErrors90, mark=\bertMark, only marks, mark size=\bertsize,  ] table [col sep=comma,
            	 y=F1BinaryMicro, 
            	 x=\multicolVal,
            	 discard if not={Embedding}{bert}, 
            	 discard if not={Defense}{100word}, 
            	 discard if not={Test}{\testVar}
            	 ]{\tablePathVal};
            	  
            	 \addplot[CErrors90, mark=\elmoMark, only marks, mark size=\elmosize,  ] table [col sep=comma,
            	 y=F1BinaryMicro, 
            	 x=\multicolVal, 
            	 discard if not={Embedding}{elmo}, 
            	 discard if not={Defense}{100word}, 
            	 discard if not={Test}{\testVar}
            	 ]{\tablePathVal};
            	 
            	 \addplot[CErrors90, mark=\gloveMark, only marks, mark size=\glovesize,  ] table [col sep=comma,
            	 y=F1BinaryMicro, 
            	 x=\multicolVal, 
            	 discard if not={Embedding}{glove}, 
            	 discard if not={Defense}{100word}, 
            	 discard if not={Test}{\testVar}
            	 ]{\tablePathVal};
                	 
            \renewcommand{\testVar}{100word} 
            	 \addplot[WErrors90, mark=\bertMark, only marks, mark size=\bertsize,  ] table [col sep=comma,
            	 y=F1BinaryMicro, 
            	 x=\multicolVal,
            	 discard if not={Embedding}{bert}, 
            	 discard if not={Defense}{100word}, 
            	 discard if not={Test}{\testVar}
            	 ]{\tablePathVal};
            	  
            	 \addplot[WErrors90, mark=\elmoMark, only marks, mark size=\elmosize,  ] table [col sep=comma,
            	 y=F1BinaryMicro, 
            	 x=\multicolVal, 
            	 discard if not={Embedding}{elmo}, 
            	 discard if not={Defense}{100word}, 
            	 discard if not={Test}{\testVar}
            	 ]{\tablePathVal};
            	 
            	 \addplot[WErrors90, mark=\gloveMark, only marks, mark size=\glovesize,  ] table [col sep=comma,
            	 y=F1BinaryMicro, 
            	 x=\multicolVal, 
            	 discard if not={Embedding}{glove}, 
            	 discard if not={Defense}{100word}, 
            	 discard if not={Test}{\testVar}
            	 ]{\tablePathVal};

\end{groupplot}
\end{tikzpicture} 
%


\centering
\footnotesize
 	\begin{tikzpicture}
 	\begin{axis}[
 	hide axis,
 	height = .75in,width=1in,
 	xmin = 0, xmax = 50, ymin = 0, ymax = 0.4,
 	legend cell align = {left}, legend columns=-1,
 	legend style = {column sep=0.05in,
 	font=\footnotesize,
 	draw=none, at = {(0.5,0.75)}},] 
 	
    \addlegendimage{empty legend}
 	\addlegendentry{\bf Test Set: }
 	\addlegendimage{Errors90, only marks, scale=2,mark=square*} 
 	\addlegendentry{\te} 
 	\addlegendimage{CErrors90, only marks, scale=2,mark=square*} 
 	\addlegendentry{ \teFullChar} 
 	\addlegendimage{WErrors90, only marks, scale=2, mark=square*} 
 	\addlegendentry{\teFullWord}

    \addlegendimage{empty legend}
 	\addlegendentry{\bf Embedding: }
 	\addlegendimage{black!50, only marks, mark size=\bertsize, mark=\bertMark} 
 	\addlegendentry{BERT} 
 	\addlegendimage{black!50, only marks, mark size=\elmosize, mark=\elmoMark} 
 	\addlegendentry{ELMo} 
 	\addlegendimage{black!50, only marks, mark size=\glovesize, mark=\gloveMark} 
 	\addlegendentry{GloVe}

 	\end{axis}
 	\end{tikzpicture}  

\vspace{-0.5\baselineskip}

    \vspace{-0.5\baselineskip}
    \caption{Binary F1 (BF1) as a function of multiclass F1 (MF1). Dashed lines indicate equal performance. 
    }
    \label{fig:high_impact}
\end{figure*}

\subsection{High Confidence Misclassifications}
Next, we examine high confidence misclassifications which are integral to understanding model behavior and the limitations faced by deceptive news detection approaches.
Figure \ref{fig:high_conf_bar} highlights error rates across test data distinguishing high confidence ($>90\%$) from lower confidence ($\le90\%$).

With the 3-way task, we observe that high confidence misclassifications account for a majority of all errors from the \trFullWord models (85.7\% of errors with the \teFullWord attack are considered high confidence). 
This is larger than the errors from any of the other models.
We also notice one exception to this finding: the \trFullWord ELMo model makes very few (less than 0.5\%) high confident incorrect predictions for the \teFullWord test set. 
With the 4-way task, we do not see the same frequency of high confidence errors although \trHalfWord displays high rates of high confidence misclassifications on BERT and ELMo models when tested with \te and \teFullChar.

Previously, we detailed stronger performance from the \EnsChar and \EnsCharWord defenses. 
As shown in Figure \ref{fig:high_conf_bar}, both ensemble defenses display the lowest (or second lowest) error rates across attacks. 
Moreover, these models exhibit sparse high confidence misclassifications when reviewing averaged confidence scores across ensembled models.
This is advantageous model behavior in a real-world setting when predicted model confidences must act as a proxy for uncertainty, and, in instances when ground truth labels are unknown, as a means to calibrate users' trust in model classifications.

\subsection{High Impact Misclassifications} 
Finally, we contrast model performance for each task and our devised binary sub-tasks (trustworthy versus deceptive for the 3-way task and satire versus not satire for the 4-way task).
Figure~\ref{fig:high_impact} demonstrates model tendencies towards high impact misclassifications across defenses, embeddings, and test sets.
A higher binary F1 score indicates fewer high impact misclassifications -- \ie more errors due to misclassifications among similar classes as compared to more errors due to misclassifications among significantly different classes.
All models exhibit higher F1 scores on the binary sub-task than the multiclass task, as would be expected since the binary task presents an ``easier'' problem with an increased random chance for correct classification.

We examine consistent trends for each test set (indicated by color) or embedding type (indicated by mark size) across defenses. 
Values plotted in the same color cluster more consistently than those plotted in the same size. 
Two defenses show the most consistency in performance across configurations.
\trHalfWord displays low performance on both the binary and multiclass formulations of the 3-way task and \trFullChar displays high performance (relative to each task) across formulations for both the 3-way and 4-way tasks, with more consistency and higher performance on the 4-way task. 
Similar to \trFullChar, \trHalfChar displays high performance across both tasks, although this defense is more consistent on the 3-way task. 
Although the \tr model is the best configuration when testing on \te, the \tr model shows much lower efficacy when tested against both attacks.
\textit{Overall, configurations using the character-based defenses result in the fewest overall high impact misclassifications.} 

Interestingly, we see that defenses are more effective at the binary sub-task for the 4-way classification (satire versus not satire) than the binary sub-task for the 3-way classification (trustworthy versus deceptive). 
Both trustworthy and deceptive news media attempt to present the information and news they share as factual, truthful content. 
In contrast, satire is distinct from other types of deceptive news as well as distinct from trustworthy news sources because it does not intend to present content as factual or accurate.
This distinction between the classes considered in the binary sub-tasks can explain the observed difference in performance.

\section{Discussion and Future Work}
 
Linguistic variation in text (adversarial or otherwise) is frequently encountered in real-world settings.
As such, we have presented extensive evaluations concerning the robustness of deception detection models to perturbed inputs.
To the best of our knowledge, we are the first to evaluate model susceptibility in regards to adversarial linguistic attacks, investigate model behavior behind high confident or high impact failures, and  present effective defensive strategies to these types of attacks.
Our comprehensive set of perturbation experiments identify key findings from not only the defender perspective (the most effective strategy of defense across multiple or combined attacks) but also the attacker perspective (the most effective method of attack) -- a focus of analysis not previously studied.
In regard to the defense viewpoint, we show that ensemble-based approaches leveraging perturbed (adversarial) and non-perturbed (original) training examples perform consistently well.
With the attack viewpoint, character-based attacks hinder performance regardless or model, defense, or task.

Our adversarial analyses have also illustrated the danger of relying on single performance metrics.
Models that achieve optimal performance on a specific task or adversarial situation may significantly under-perform with slight alterations in scope or context.  
For example, although the \EnsChar and \EnsCharWord models saw second best performance on either classification task, they outperformed the ``best models'' when considering all possible attacks.
The models with the highest overall performance were also not consistently found to have the lowest high confidence or high impact misclassifications -- an important consideration if a model is being considered for use on live platforms where decisions can significantly impact users.

The results highlighted in this work provide justification for enhanced development and analysis of deception detection models. 
Although we rely on a consistent model architecture in order to make equitable comparisons across tasks and datasets, the evaluation framework we present can be replicated with additional models, complex architectures, and variants in test data.
This work relies on uniform perturbation attacks as opposed to strategic perturbation strategies that target specific substrings -- such as pseudonymous terms, phrases, or monikers. 
Subsequent experiments will investigate more complex strategic attacks and their ability to evade or confuse deception detection models.

\section*{Acknowledgements}
 This research was supported by the Laboratory Directed Research and Development Program at Pacific Northwest National Laboratory, a multi-program national laboratory operated by Battelle for the U.S. Department of Energy.
 
\bibliographystyle{acl_natbib}

\end{document}